\documentclass{article} % For LaTeX2e
\usepackage{iclr2026_conference,times}

\usepackage{amsmath,amsfonts,bm}

\def\eqref#1{equation~\ref{#1}}
\def\1{\bm{1}}

\DeclareMathAlphabet{\mathsfit}{\encodingdefault}{\sfdefault}{m}{sl}
\SetMathAlphabet{\mathsfit}{bold}{\encodingdefault}{\sfdefault}{bx}{n}

\usepackage[T1]{fontenc}
\usepackage[utf8]{inputenc}
\usepackage{hyperref}
\usepackage{url}
\usepackage{cleveref}
\usepackage{subcaption}  % 推荐使用 subcaption 而不是 subfig 或 subfigure
\usepackage{graphicx}
\usepackage{pifont}
\usepackage{amsthm,amsmath,amssymb}
\usepackage{multirow} 
\usepackage{tcolorbox}
\tcbuselibrary{skins, breakable}
\usepackage{hyperref}
\usepackage{enumitem}
\usepackage{multicol}
\usepackage{booktabs} 
\usepackage{fontawesome}
\usepackage{array}
\usepackage{colortbl} % 用于设置表格颜色
\usepackage{xcolor} % 提供颜色支持
\definecolor{bgblue}{RGB}{218, 232, 252}
\definecolor{bgpurple}{RGB}{225, 213, 231}

\title{AutoTool: Automatic Scaling of Tool-Use Capabilities in RL via Decoupled Entropy Constraints}
\author{Yirong Zeng\textsuperscript{1}, Xiao Ding\textsuperscript{1*}, Yufei Liu\textsuperscript{2}, Yuxian Wang\textsuperscript{3}, Qunyao Du\textsuperscript{1}, Yutai Hou\textsuperscript{3*} \\
Wu Ning\textsuperscript{3}, Haonan Song\textsuperscript{3}, Duyu Tang\textsuperscript{3}, Dandan Tu\textsuperscript{3}, Bing Qin\textsuperscript{1}, Ting Liu\textsuperscript{1} \\
\AND
\textsuperscript{1}Harbin Institute of Technology, SCIR Lib\\ 
\textsuperscript{2}Peking University\\
\textsuperscript{3}{Huawei Technologies Ltd.} \\
\AND
\textsuperscript{*}Corresponding authors\\
\texttt{{\{yrzeng,xding}\}@ir.hit.edu.cn, houyutai@huawei.com} \\
}

\iclrfinalcopy % Uncomment for camera-ready version, but NOT for submission.
\begin{document}

\maketitle

\begin{abstract}

Tool use represents a critical capability for AI agents, with recent advances focusing on leveraging reinforcement learning (RL) to scale up the explicit reasoning process to achieve better performance.
However, there are some key challenges for tool use in current RL-based scaling approaches: 
(a) direct RL training often struggles to scale up thinking length sufficiently to solve complex problems, 
and (b) scaled-up models tend to overthink simpler problems, resulting in substantial token inefficiency.
To address these challenges, we propose a novel training paradigm that first employs warm-up supervised fine-tuning to help models distinguish between simple and complex problems, followed by RL that enable models to automatically determine appropriate reasoning trajectories. 
Furthermore, to tackle the issue of automatic thinking-length scaling, we discover that entropy-based optimization objectives effectively maintain model diversity while successfully unlocking the model's scaling capabilities.
Based on this insight, we introduce an entropy-based long-short reasoning fusion RL strategy. 
Our experiments on three benchmarks demonstrate that model successfully achieves auto-scaling for efficient tool use, achieving significant 9.8\% accuracy improvements while reducing computational overhead by \textasciitilde81\%.

% Reinforcement Learning with verifiable rewards has emerged as a robust training paradigm by scaling test-time reasoning. 
% Reasoning collapse, where short trajectories dominate, causing the model’s reasoning behavior to shrink from long to short trajectories with low entropy, which hinders effective scaling.
% 复杂问题长度向上冲一下后，下降。简单问题真实下降 （基于输入和输入长度衡量简单和复杂）
% 不能让短路径因“快而准”垄断优势，必须保护长路径的探索空间。 
% GRPO 的组内相对优势机制会放大“快而准”的短路径优势，导致模型“懒惰化”：所有问题都倾向于走短路径 （全局熵坍塌）
% leading to model laziness3

% 重解决问题的 介绍问题动机(1->2)
% 主要方法创新的文章重点说方法(只说关键逻辑，highlight 关键创新和解决策略，不涉及具体细节方案, 即流水帐不行，2-3句话)
% 实验结果吸引人 多介绍实验结果（1->2句）
% 写实验分析时，首段第一句话，先写结论，后先实验设置及分析。
\end{abstract}

\section{Introduction}
% P1: Agent tool use 的意义
Integrating agentic large language models (LLMs) with external tools has emerged as a pivotal advancement, and has become a defining feature of advanced agentic models \citep{OpenAIDeepResearch2025,KimiResearcher2024}.
It significantly enhances a model's ability to address complex tasks \citep{qu2025tool,wang2024executable}, and opens up many practical uses across different fields. 
For example, it supports the automation of reasoning tasks \citep{jin2025search,li2025torl}, and enables agent applications \citep{ragen,Agent-R1}.
Therefore, research on agentic tool use represents a critical pathway toward artificial general intelligence.
In this task, models respond to queries by dynamically selecting and invoking relevant tools from an available pool.
% In this paper, tools are used interchangeably with APIs, functions, and plugins.

% P2: 当前工具学习的方法 + 问题
% Test Time Scale (TTS, \citep{deepseekai2025r1,muennighoff2025s1}) is defined as a training strategy that scales up a model’s response length to unlock its intrinsic reasoning capabilities, and it is closely associated with Reinforcement Learning with Verifiable Rewards (RLVR) in LLMs training.
Test-time scaling (TTS) is a approach to language modeling that uses extra test-time compute to improve performance \citep{muennighoff2025s1}.
Currently, scaling up a model’s explicit reasoning length via Reinforcement Learning with Verifiable Rewards (RLVR) is a effective to achieve TTS.
% Recently, Reinforcement Learning with Verifiable Rewards (RLVR) has demonstrated strong reasoning abilities via test time scale (TTS, \citep{deepseekai2025r1,muennighoff2025s1}), a strategy that scales up model response length during training to unlock intrinsic reasoning. 
Compared to the prevalent Supervised Fine-Tuning (SFT) approach, which imitates reasoning patterns from labeled high-quality examples to teach models using tools \citep{liu2024toolace,zhang2024xlam},
RLVR better fosters intrinsic reasoning, rather than making models memorize training trajectories \citep{chen2025acebench}.
% However, SFT often leads models to memorize training trajectories instead of developing robust, intrinsic reasoning capabilities 
It has demonstrated robustness in mathematics \citep{shao2024deepseekmath} and coding \citep{pan2025metaspatial}, while also driving a paradigm shift from SFT to RL in LLM training.
Thus, exploring suitable scaling strategies is critical to advance effective agentic tool-use.

\begin{figure*}[t]
    \small
    \centering
    \begin{subfigure}[t]{0.32\linewidth}
        \centering
        \includegraphics[width=0.99\linewidth]{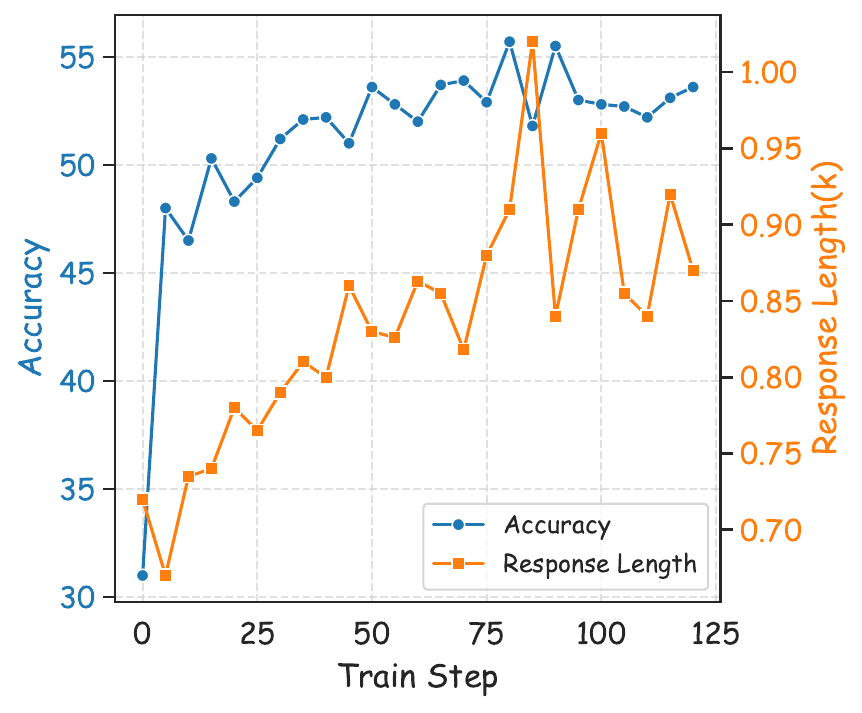}
        \caption{ direct RL training in Math}
        \label{fig:exp1a}
    \end{subfigure}
    % \hfill
    \begin{subfigure}[t]{0.32\linewidth}
        \centering
        \includegraphics[width=0.99\linewidth]{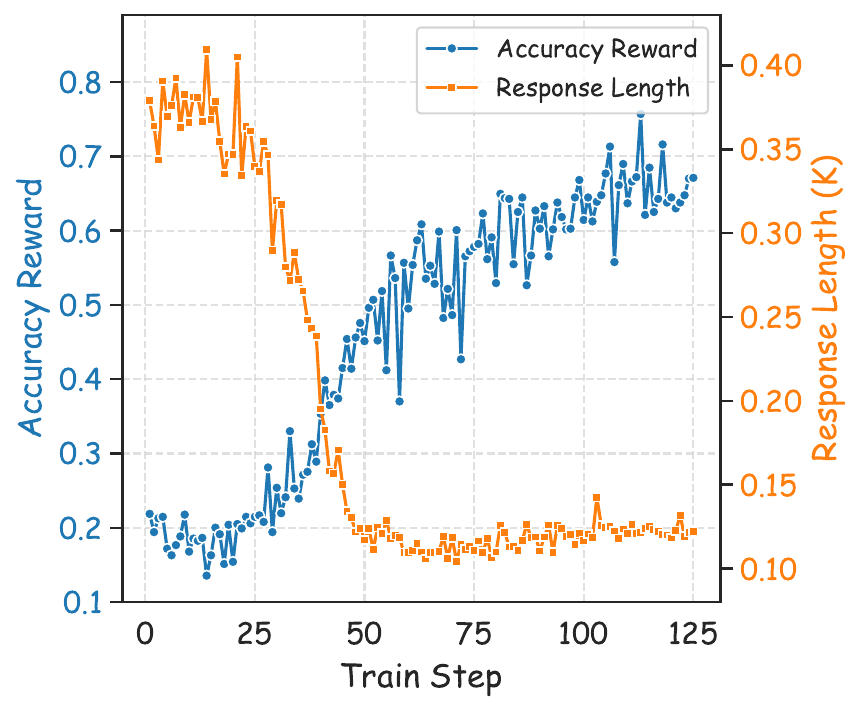}
        \caption{direct RL training in tool use}
        \label{fig:exp1b}
    \end{subfigure}
    \begin{subfigure}[t]{0.32\linewidth}
        \centering
        \includegraphics[width=0.99\linewidth]{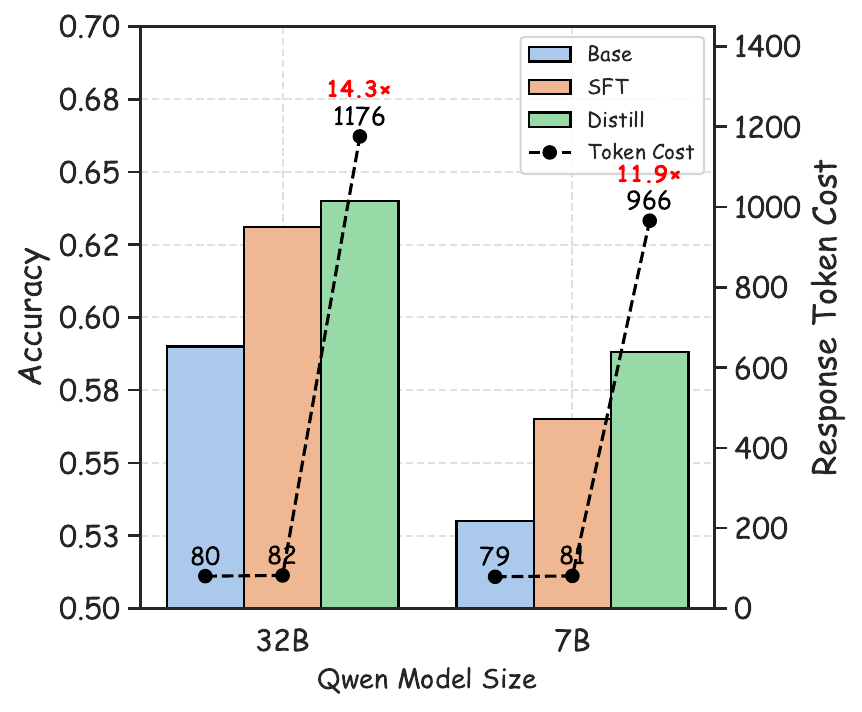}
        \caption{Cost in scaled-up models}
        \label{fig:exp1c}
    \end{subfigure}
  \caption{
    The training paradigms for TTS in tool-use:
    (a) direct RL enables scaling up response length as accuracy improves in mathematical tasks; but (b) it fails to scale in tool-use tasks, where reasoning collapses into short trajectories; (c) scaled-up models (e.g., distillation models) incur significant token costs, as they require lengthy reasoning trajectories for all queries.
    % (a) It incurs significant token costs due to lengthy reasoning trajectories for all queries.
    % (b) Training with RL, where a performance-length decoupling emerges, reasoning collapse into short trajectories, undermining advanced tool use.
    }
  \label{fig:motivation}
\end{figure*}

% P3 问题分析
% driven by the dominant policy advantage of short-reasoning trajectories.
To this end, we pre-analyze the training paradigm for TTS in tool use, as shown in Figure \ref{fig:motivation}.
Under direct RL training, contrary to mathematical tasks where response length scales with improving accuracy, we observe that models suffer from \textit{reasoning collapse} in tool use, a phenomenon where models fail to sufficiently extend thinking\footnote{
In this paper, the terms \textit{thinking} and \textit{long reasoning} are used interchangeably, both referring to responses that contain an explicit reasoning process.
}  length to solve complex problems. 
More importantly, we find many tool-use problems can be solved with short reasoning trajectories, yet scaled-up models generate excessively long trajectories that cause unnecessary resource consumption. 
Therefore, an adaptive model that dynamically integrates short and long reasoning is highly desirable. 
% Recently, several works have explored automatic scaling , i.e., enabling models to adaptively select the optimal reasoning mode based on problem difficulty \citep{fang2025thinkless,zhang2025adaptthink,huang2025adactrl,wang2025think}.

% \begin{figure*}[th]
%   \centering
%   \includegraphics[width=0.6\linewidth]{./decoupled.pdf}
%   \caption{
%   An illustration comparing decoupled adaptive entropy constraints with native RLVR in agentic tool-use.
%     $\mathcal{L}$ is the policy loss of RLVR, and $\mathcal{\beta}$ represents the entropy coefficient.
%      Our method decouples different reasoning modes by applying differentiated entropy constraints: entropy strength is adjusted above a target entropy for long trajectories, while remaining fixed for short trajectories.
%     }
%   \label{fig:overview}
% \end{figure*}
% P4 实验方法 + 结果 + 贡献

More analysis in Section \ref{sec:pre_study} reveal that low entropy, which quantifies a model’s response certainty and exploration capability, leads to insufficient reasoning length, limiting the robustness of LLMs in tackling hard problems. 
These findings motivate our proposed method, which decouples long and short reasoning to prevent dominance interference while incorporating entropy constraints to enable long-trajectory reasoning.
Therefore, we propose a decoupled adaptive entropy constraint strategy for RL. 
It first performs warm-up using a constructed mixed dataset of long and short reasoning trajectories to perceive data difficulty. 
The strategy decouples the policy loss between short and long reasoning, 
then applies varying entropy constraint strengths to regulate the thinking mode while maintaining concise responses for simple problems. 
This enables differentiated exploration control across reasoning modes, with adjusted entropy strength set above a target entropy in long reasoning to preserve exploration capacity.
% 可控的推理模式 by 预拼接 reaoning pattern tokens in input tokens
% Additionally, we constructed a general-purpose tool-use dataset with multi-turn interaction based on public open-source data, to support RL training.

Experiments on three benchmarks show:
(1) Our \textasciitilde7B model leads at comparable size models (e.g., +11.95\% compare to SFT-model).
(2) Beyond performance, our auto-scaling model boosts accuracy by 9.8\% compared to the distilled model and cuts inference token cost by \textasciitilde81\%.
Notably, our model’s thinking rate reaches 45\% in complex scenarios but 0\% in simple ones. 
Moreover, visualizations of the training process confirm that our approach generates concise responses for simple cases while extending reasoning trajectories by $5 \times$ for complex questions.
This contrast demonstrates that the model has learned to automatically adjust test-time scales based on sample difficulty, ultimately supporting improved inference efficiency.

\section{Preliminary Study}
\label{sec:pre_study}
In this section, we present extensive experiments to highlight the challenges of achieving test-time scaling for agentic tool use, and thereby motivate our method.

\subsection{Task Overview}
 In agentic tool use, the LLM receives a user query \( q \) along with candidate tools, represented as \( \mathcal{T} = \{ {t}_0, {t}_1, \dots, {t}_{|\mathcal{T}|} \} \). 
 The purpose of LLM is to fulfill the user’s intent by executing a specific sequence of tools. 
 We formalize this decision-making process as \( y \sim \pi(y \mid q, s, \mathcal{T} ) \), where \( \pi(\cdot) \) represents the policy model, \( s \) denotes the task state (e.g., historical context ), and \( y \) represents the actions taken by the model, such as selecting or executing tool calls. 
A review of related work is provided in Appendix~\ref{sec:relatedwork}.
 % Each decision may update the state \( s_i \), guiding subsequent decisions.

\subsection{Training Paradigms Analysis}

We analyze training paradigms for scaling reasoning process in tool use, including RL training and SFT with distillation, using Qwen2.5-series models to conduct training on the public \textit{ToolACE} dataset \citep{liu2024toolace} and evaluation via \textit{BFCL} \citep{bfclv3} (details in Section \ref{exp_setup}).

\noindent
(1) For direct RL, we applied RL (specifically GRPO \citep{shao2024deepseekmath}). 
As shown in Figure~\ref{fig:exp1b}, we observe a divergence between the model's performance and response length: as training steps increased, performance improved while response length decreased sharply.
This indicates the reasoning patterns collapsed into short reasoning trajectories, failing to scale-up in test-time. 
This result contradicts widely accepted findings from training on complex reasoning tasks (e.g., mathematics) \citep{openr1, zeng2025simplerl2}, as shown in Figure~\ref{fig:exp1a}, where we adopt experimental results from \citet{zeng2025simplerl}.
The evaluation results presented in Table \ref{tab:pre-exp} indicate that performance in complex tool-use scenarios (e.g., Multi-Turn) decreased noticeably compared to the distilled SFT model.
This indicates that the reasoning collapse phenomenon limits the model's robust performance on complex problems.

\noindent
(2) For SFT with distillation, we conducted base SFT and distillation from reasoning LLM \citep{ds0528}, respectively.
As shown in Figure \ref{fig:exp1c}, distilled models showed no noticeable accuracy gain over the base SFT, but increased output token costs by more than $10 \times$.
This suggests that many agentic tool-use problems can be solved with short reasoning trajectory while excessive long trajectory leads to unnecessary resource consumption.

\begin{figure*}[t]
    \centering
  \includegraphics[width=0.99\linewidth]{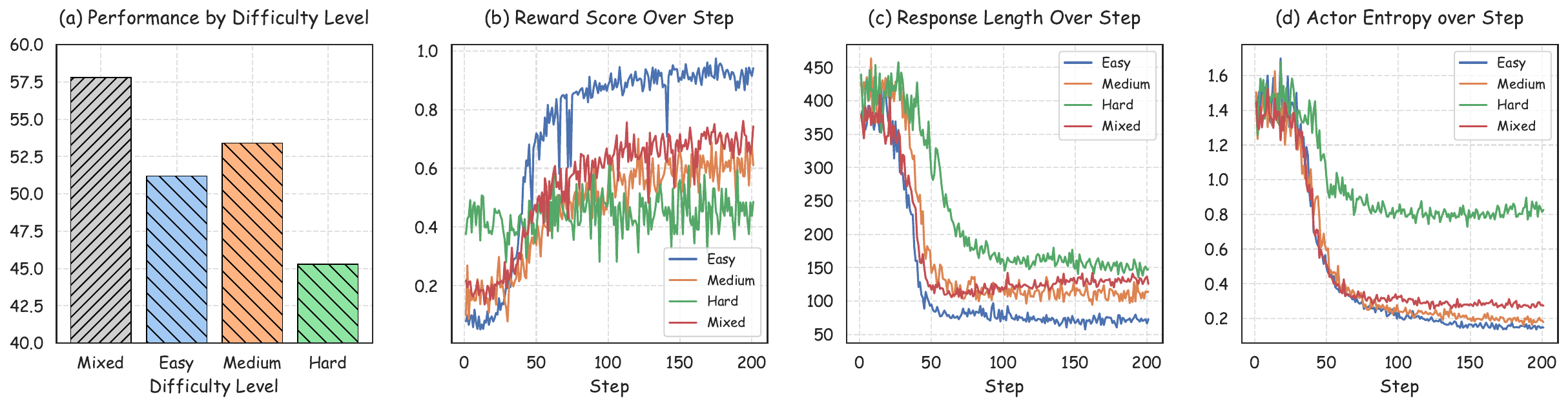}
  \caption{Impact of difficulty distributions. 
      \textit{Easy} and \textit{Medium} converged successfully, while \textit{Hard} failed (a, b). 
      However, collapse occurred across all three subsets (c), with the same trend observed in entropy (d).
      % This indicates data distribution 与collapse无关联，而entropy有强正关联。
      This indicates that data distribution has no correlation with collapse, whereas low entropy exhibits a strong positive correlation.
      % is not the main factor influencing reasoning pattern collapse.
    }
  \label{fig:pre-exp2}
\end{figure*}

% In summary, our analysis identifies critical flaws in existing TTS paradigms: distilled SFT markedly increases token costs, while RL-scaled models suffer from reasoning collapse.
% To address this, we aim to develop an adaptive model that dynamically combines short and long reasoning, boosting both complex tool-use capability and computational efficiency.
% A key challenge is mitigating reasoning pattern collapse during RL training for agentic tool use.

\subsection{Pre-study on Reasoning Pattern Collapse.}
To investigate the causes of reasoning pattern collapse, we conducted an in-depth analysis of data difficulty distribution and information entropy.

\subsubsection{Data Distribution}
\label{sec:data_dist}
Intuitively, we hypothesized that the sample difficulty distribution might exert a critical influence.
Guided by this hypothesis, we used a base model to perform 8 rounds of reasoning on the training data and calculated pass@8.
The resulting distribution (shown in Appendix ~\ref{sec:dataset}) reveals that \textbf{easy} samples (with $8/8$ correct inferences) and \textbf{hard} samples (with $0/8$ correct inferences) accounted for 47\% and 31.8\% of the dataset, respectively, while \textbf{medium} samples made up a smaller proportion.

We then conducted separate RL training runs on these three subsets, evaluated the resulting models, and reported their training dynamics in Figure~\ref{fig:pre-exp2}.
Notably, reasoning pattern collapse persisted across all three subsets: after an initial exploration phase, easy samples led to rapid convergence, intermediate samples resulted in slower convergence, and hard samples showed no convergence. 
These findings indicate that the sample difficulty distribution can slightly reduce convergence speed, but no correlation with collapse has been observed.

% Additionally, as reasoning pattern collapse occurs on easy data, the other two models also exhibit this phenomenon sequentially, eventually converging to short reasoning paths. 
% This observation suggests that the intra-group relative advantage mechanism of GRPO amplifies the advantages of "fast and accurate" short paths, allowing such paths to dominate. 
% Therefore, we hypothesize that decoupling policy loss by sample difficulty, preventing easy example from overwhelming the gradient signal, can mitigate the monopolization of short paths.

The actor model's information entropy quantifies its exploration capability during training.
As shown in Figure \ref{fig:pre-exp2}d, entropy decreases rapidly, with dynamics closely aligning with reasoning pattern collapse.
Additionally, comparing three subsets, the final converged entropy increases sequentially from simple to complex samples.
This reveals: simple problems elicit high certainty in short reasoning (perceiving extended exploration risks suboptimal solutions); complex problems face inherent challenges, with short reasoning advantage dominance further discouraging exploration and driving default to brief responses.
This finding demonstrates a strong positive correlation between low entropy and collapse.
% thus, preventing excessively low entropy may mitigate the collapse
% 这说明在简单问题上，模型认为不用多想，想多了可能越来越偏，因此以高确定性趋向短回答。
% 复杂问题本身有挑战，加上优势被垄断， 摆烂不想探索了，因此变成短回答。
% 因此，阻止信息熵过低，是一个可能的解决策略。

\begin{table*}[t]
    \centering
    \small
    \caption{Evaluation results on the BFCL benchmark, which includes three sub-metrics: Non-live, Live, and Multi-Turn (including multi-turn and long-context tool-use scenarios). $l$ denotes target response length in response constraint, and $\beta$ denotes coefficients of entropy constraint.
    % \textbf{Bold} for best performance and \underline{underline} for best performance in the other types.
    }
    \label{tab:pre-exp}
    \begin{tabular}{l c| ccc|c}
        \toprule
         Model & Think? & Non-Live & Live & Multi-Turn & {Overall Acc} \\
        \midrule
        Base LLM & \ding{55} & 86.46 & 67.44 & 7.62 & 53.69 \\ 
        \hspace{0.2cm} w/ SFT & \ding{55} & {86.65} & 75.11 & 6.75 & 56.90 \\
        \hspace{0.2cm} w/ distilled SFT & \ding{51} & 87.35 & 79.59 & 16.95 & 59.23 \\
        \hspace{0.2cm} w/ GRPO & \ding{51} & 87.06 & 78.22 & \hspace{0.2cm}8.38\textsubscript{\textcolor{red!70!black}{↓8.57}} & 57.81 \\
        \midrule
        w/ length constraint \\
            \hspace{0.4cm} $+l = 100$ & \ding{51} & 87.30 & 71.23 & 7.92 & 55.37 \\
            \hspace{0.4cm} $+l = 50$ & \ding{51} & 87.76 & 78.43 & 8.78 & 58.12 \\
            \hspace{0.4cm} $+l = 10$ & \ding{51} & 89.76 & 77.33 & 8.89 & 58.27 \\
        \multicolumn{6}{l}{w/ entropy constraint} \\
        \hspace{0.4cm} $+\beta = 1e{-2}$ & \ding{51} & 87.47 & 79.13 & 9.48 & 59.33 \\
        \hspace{0.4cm} $+\beta = 5e{-2}$ & \ding{51} & 87.21 & 77.96 & 10.02 & 58.91 \\
        \hspace{0.4cm} $+\beta = 1e{-1}$ & \ding{51} & 88.32 & 80.42 & \hspace{0.2cm}15.86\textsubscript{\textcolor{green!70!black}{↑7.48}} & 61.86 \\
        \bottomrule
    \end{tabular}
\end{table*}

\begin{figure*}[t]
    \centering
    \small
      \begin{subfigure}[t]{0.45\linewidth}
        \centering
        \includegraphics[width=0.99\linewidth]{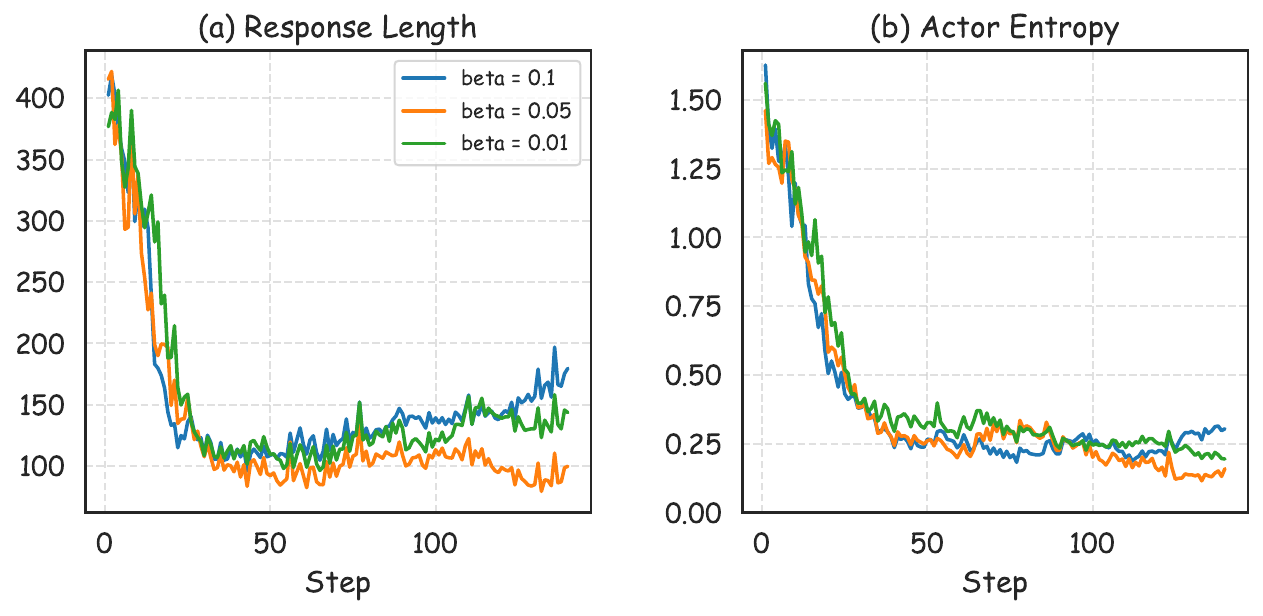}
        \caption{RL with entropy constraint}
        \label{fig:exp1a}
    \end{subfigure}
    % \vspace{0.3em}
    \hfill
    \begin{subfigure}[t]{0.45\linewidth}
        \centering
        \includegraphics[width=0.99\linewidth]{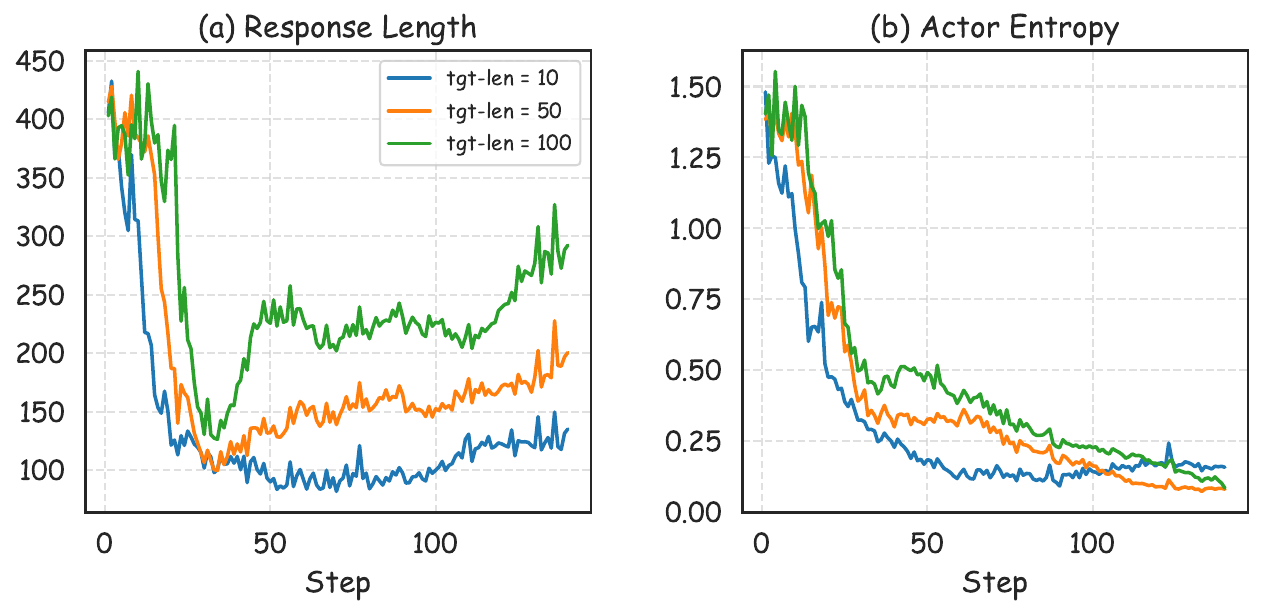}
        \caption{RL with length penalty}
        \label{fig:exp1b}
    \end{subfigure}
  \caption{
    Training dynamics visualized for entropy constraint (a) and length penalty (b).
    The entropy constraint partially increases response length, yet the length penalty does not mitigate low entropy.
  }
  \label{fig:pre-exp32}
\end{figure*}

\subsubsection{Information Entropy Constraints}
To explore reasoning collapse-entropy connections, we incorporated entropy constraints into the policy loss function. 
Inspired by \citet{he2025skywork}, we designed a mechanism to maintain the entropy ($e$) at a reasonable level throughout training.
The entropy loss is defined as: 
\begin{equation}
    loss_k^{e} = \beta \cdot \mathbb{I}{\{e_k \leq \texttt{tgt-ent}\}}
\end{equation}
where $k$ is the training step, $\beta$ is the coefficient, and we set \texttt{tgt-ent}=0.1.
Notably, the entropy loss is only activated when $e_k \leq$ \texttt{tgt-ent}, ensuring the model’s entropy remains lower-bounded by the target value.
For comparative purposes, we also implemented a short-response penalty by configuring the reward function to penalize response below the target length $l$.
The evaluation results (presented in Table~\ref{tab:pre-exp}) show that the length constraint did not improve the model’s test performance. 
In contrast, the entropy constraint yielded partial performance gains (visualizing the training process in Figure~\ref{fig:pre-exp32}). 
However, the effectiveness is highly sensitive to $\beta$: 
in multi-turn, the maximum positive gain was achieved when $\beta=1e-1$, whereas results for other \(\beta\) are comparable to \textit{w/ length constraints}.
This sensitivity highlights the difficulty of pre-selecting an optimal entropy coefficient, indicating that dynamically adjusting $\beta$ during training is necessary.

Therefore, we propose a novel strategy: decoupled adaptive entropy constraints. 
It decouples entropy constraint for short and long reasoning and adaptively tunes the entropy coefficient, addressing both the collapse caused by low entropy and the sensitivity of static coefficients.

\section{Methodology}
\label{sec:method}
\begin{figure*}[t]
    \centering
  \includegraphics[width=0.85\linewidth]{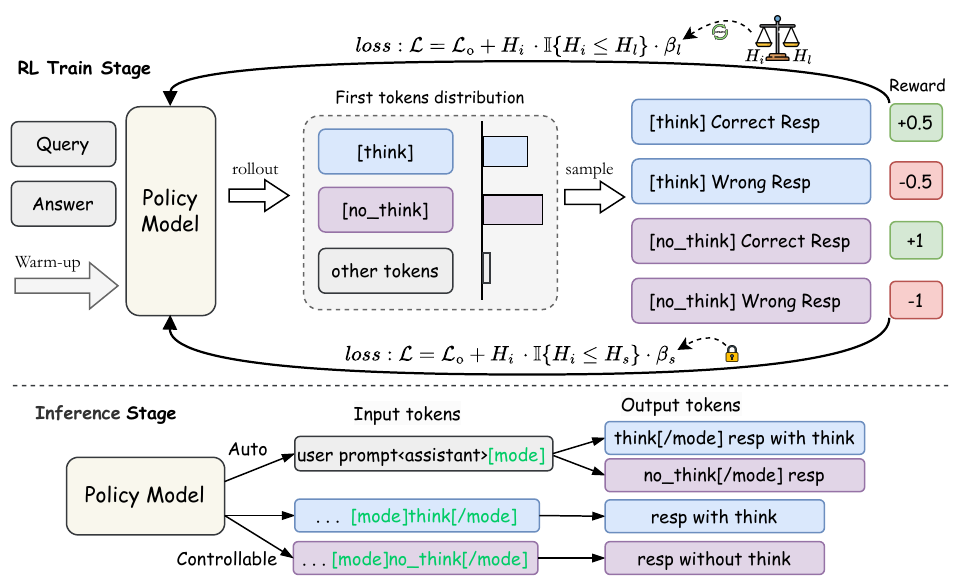}
  \caption{The overview of decoupled adaptive entropy constraint.
  It achieves automatic scaling by decoupling different reasoning modes through the application of differentiated entropy constraints. 
  Adaptive entropy constraint strength for long reasoning.
  During the inference, the model can automatically or controllably switch inference modes by pre-pending a \textcolor{green!80!black}{response prefix} in Input tokens. }
  % in RL train stage, \fcolorbox{gray}{bgblue}{\textcolor{black}{[think]}} and \fcolorbox{gray}{bgpurple}{\textcolor{black} {[no\_think]}} denotes \texttt{[mode]think[/mode]} and \texttt{[mode]no\_think[/mode]}, respectively.
  \label{fig:main}
\end{figure*}

% In this section, we provide a detailed introduction to our method, as shown in Figure~\ref{fig:main}.
Our method first performs warm-up SFT to perceive sample difficulty (details in \ref{sec:dataprepare}), followed by RLVR training with decoupled adaptive entropy constraints, as shown in Figure~\ref{fig:main}.
% We first describe the , followed by an overview of our method, as shown in Figure~\ref{fig:main}.

% In contrast, $\beta_s$ remains fixed during training, resulting in decoupled entropy control between short and long paths.

% Data Preparation and Warm-up
\subsection{Data Preparation and Warm-up}
\label{sec:dataprepare}
To support robust general-purpose tool use via RL, we constructed a mixed dataset covering diverse tool-use scenarios from public sources: ToolACE \citep{liu2024toolace}, xLAM \citep{zhang2024xlam,prabhakar2025apigen}, Hermes Function-Calling \citep{Hermes-Function-Calling-Dataset-V1}.
More details are provided in Appendix \ref{sec:dataset}.
To create a balanced dataset encompassing diverse complexity levels and tool usage scenarios, we randomly downsampled the raw data.
Moreover, we adopted the following strategies to develop a \underline{Pub}lic agentic \underline{Tool}-use dataset (\textbf{PubTool}), as presented in Table~\ref{tab:data_stats}.

% \textbf{Data Distillation for Warm-up SFT.}
\textbf{Warm-up Training.}
To help the model initially perceive data difficulty, we propose SFT for warm-up training by mixing long and short reasoning data.
To construct such mixed thinking data, we performed multiple inferences (calculating pass@8) on the training data using Qwen2.5-7B-Instruct (no-thinking model) and Qwen3-32B (thinking model), respectively. For each response turn, we adopted the ground truth as the label if the no-thinking model's output was correct; otherwise, we adopted the thinking model's answer with explicit long reasoning if it was correct.
More Details of data preparation are shown in Appendix \S\ref{sec:dataset}.
% To balance the thinking rate, we downsampled short responses, with results shown in Table \ref{tab:data_stats} (Appendix \S\ref{sec:dataset} for more data prepare details). 
We design an auto-thinking template (details in Appendix \S\ref{sec:auto_prompt}) to enable the model to select reasoning modes based on data difficulty. 
% To enable this behavior, we propose an auto-thinking prompt for training and inference process, details in Appendix \S~\ref{sec:auto_prompt}.
Finally, we conducted SFT on the base model for warm-up, preparing for subsequent RL scaling.

\begin{table}[th]
  \small
  \centering
  \caption{Data statistics of \textbf{PubTool} in data collection and construction. 
    Subscript text in the SFT data table indicates the thinking rate in all turns.
  }
  \label{tab:data_stats}
  \begin{tabular}{@{}l|ccc@{}}
    \toprule
          & ToolACE & xLAM  & Hermes Function-Calling \\
    \midrule
    Raw Data   & 11.3k     & 65k   & 7.1k     \\
    Downsampled    & 11.3k     & 15k   & 7.1k     \\
    \midrule
     & \multicolumn{3}{c}{\textbf{PubTool}} \\ \midrule
    \multirow{2}{*}{Processed}      & \multicolumn{2}{c}{SFT data} & RL data \\
                  & \multicolumn{2}{c}{8.2k$\scriptstyle{(9.2\%)}$}    &  7k \\
    \bottomrule
  \end{tabular}
\end{table}

\textbf{Quality Refinement for RL data}. 
To efficiently support auto-scaling RL training, we employed the following data enhancement strategies:
First, from data distribution analysis in Section \S \ref{sec:data_dist}, we observed that the original dataset was dominated by overly simple and excessively difficult samples. 
Overly simple samples offer limited value for RL training and lack generalization, while overly difficult samples either exceed model capabilities or contain noise. 
We therefore randomly removed half of both simple and difficult samples to balance the dataset distribution.
Additionally, inspired by \citet{li2025limr}, we prioritized training samples based on their alignment with model learning trajectories.
Specifically, we performed multi-epoch GRPO training on all training data, computed changes in their reward scores, and calculated each sample's variance relative to the average reward. 
Lower variance indicated higher alignment. 
Through these processes, we downsampled the RL dataset from 21k to 7k samples.
% , which facilitates efficient resource utilization and scalable implementation.
For a detailed analysis of its effects, please refer to Appendix \ref{sec:dataset}.

\subsection{Decoupled Adaptive Entropy Constraints}
\label{sec:method}
To enable automatic scaling in agentic tool use, we propose a \textit{decoupled adaptive entropy constraints} strategy for RLVR.
 % In addition, we consider the clip-higher trick proposed in [34] as another means of entropy control. , DAPO \citep{yu2025dapo} 
The objective policy loss integrates the surrogate objective from native RLVR (e.g., GRPO) with 
a mechanism that: 
(1) {decouples} entropy regulation between short and long trajectories; 
(2) {adaptively adjusts} the entropy strength in long reasoning trajectories to preserve exploration capacity.

Specifically,
let $ \pi_\theta $ be the policy, $ H_i = -\mathbb{E}_{a \sim \pi_\theta(\cdot|s_i)}[\log \pi_\theta(a|s_i)] $ is the entropy at step $ i $, and $ m_i \in \{0,1\} $ an indicator variable:
 % is a binary indicator variable that 
 equals 1 if the action step is a short trajectory and 0 if it is a long trajectory.
We apply decoupled entropy constraints based on policy model's response trajectory type:
(1) $ \beta_s $: fixed coefficient for short paths (to prevent excessive exploration),
(2) $ \beta_l $: adaptive coefficient for long paths (learned dynamically).

% The sample-level policy loss is defined as:
% \begin{equation}
%     \mathcal{L}_{\text{p}} = \frac{1}{N} \sum_{i=1}^{N} \left[ 
%     -\min\left( \rho_i \hat{A}_i,\ \text{clip}\left(\rho_i, 1-\epsilon, 1+\epsilon\right) \cdot \hat{A}_i \right) 
%     - \mathcal{L}_{\text{ent}}^{(i)} \right],
% \end{equation}
% where the per-sample entropy regularization term is:
% \begin{equation}
%     \mathcal{L}_{\text{ent}}^{(i)} = 
%     \beta_s m_i \mathbb{I}\{H_i \leq H_S\} H_i + \beta_l (1 - m_i) \mathbb{I}\{H_i \leq H_L\} H_i.
% \end{equation}
The sample-level policy loss is defined as:

\begin{equation}
    \beta_i = \beta_s \cdot m_i \cdot \mathbb{I}\{H_i \leq H_s\} + \beta_l \cdot (1 - m_i) \cdot \mathbb{I}\{H_i \leq H_l\},
\end{equation}

\begin{equation}
    \mathcal{L}_{\text{p}} = \frac{1}{N} \sum_{i=1}^{N} \left[ 
    -\min\left( \rho_i \hat{A}_i,\ \text{clip}(\rho_i, 1-\epsilon, 1+\epsilon) \cdot \hat{A}_i \right) 
    - \beta_i H_i \right],
\end{equation}
where $\beta_i $ adapts the entropy penalty per sample, $ \rho_i = \pi_\theta(a_i|s_i) / \pi_{\theta_{\text{old}}}(a_i|s_i) $, $H_l$ and $H_s$ denote target entropy of long reasoning and short reasoning, and $ \hat{A}_i $ is the estimated advantage based on reward scores in Section \ref{sec:reward}. 
The key design is the \textit{decoupling} of entropy weights via $ m_i $, enabling distinct regularization strategies.

\textbf{Adaptive Entropy Coefficient Loss.}
Entropy regularization is highly sensitive to the choice of coefficient, making it difficult to select an optimal coefficient in advance.
This motivates a dynamic adjustment of the entropy loss coefficient. 
To automatically adjust the entropy strength for long trajectories, we introduce an adaptive loss that updates $ \beta_l $ based on the deviation of actual entropy from a target level.
The loss is computed only on steps belonging to long trajectories ($m_i = 0$) and is defined as:
\begin{equation}
    \small
    \mathcal{L}_{\beta}^l = \frac{1}{\sum_j (1 - m_j)} \sum_{i=1}^{N} (1 - m_i) \cdot \beta_l \cdot (H_i - H_l),
\end{equation}
where $H_l$ is a predefined target entropy.
The coefficient $\beta_l$ is updated by minimizing $\mathcal{L}_{\beta}^L$: if $H_i < H_l$,  $\beta_l$  increases to encourage exploration; 
if $H_i > H_l$, it decreases to suppress excessive randomness.
In contrast, $\beta_s$ remains fixed during training.
% Note that $\beta_s$ is kept fixed separately, while $\beta_l$ is updated via this loss.

\subsection{Auto Thinking Reward Module}
\label{sec:reward}
In this module, the model's output is evaluated using a rule-based reward \citep{deepseekai2025r1,meng2025mm} to compute the estimated advantage for the objective loss $\mathcal{L}_{\text{p}}$.
Specifically, for each question $q$, model generates $G$ completions $\{o_1, o_2, \dots, o_G\}$ using $\pi_{\theta_{\text{old}}}$.
This reward module combines format and answer rewards to score each completion.

\textbf{Format Reward.} 
The format reward $\mathcal{R}_{\text{format}}(o_i) \in \{0, 1\}$ evaluates whether the output adheres to the required structural template.
% This binary scoring ensures the model respects the system’s structural expectations.
We define two valid reasoning modes: \textit{think} and \textit{no-think}, each with strict syntactic constraints:

\begin{center}
\begin{tcolorbox}[
    colback=blue!8!white,       % 背景色
    colframe=blue!40!black,% 边框色
    boxrule=0.8pt,
    arc=4pt,              % 圆角
    boxsep=1pt,
    left=4pt, right=1pt, top=3pt, bottom=3pt,
    fontupper=\ttfamily\small,
    breakable,
    width=0.88\linewidth
]
[mode]think[/mode][think]reasoning process here[/think]answer

[mode]no\_think[/mode][no\_think]\textbackslash n[/no\_think]answer
\end{tcolorbox}
\end{center}

This design encourages explicit reasoning for complex problems via the \textit{think} mode, while allowing direct generation for simple queries via \textit{no-think}, reducing computational overhead. 
% By enforcing format compliance, the reward promotes efficient behavior: simple problems are answered concisely, while complex ones are addressed with structured, step-by-step reasoning, cultivating the model's intrinsic reasoning capability when needed.
% To enable this behavior, we propose an auto-thinking prompt for training and inference process, details in Appendix \S~\ref{sec:auto_prompt}.
% We define an auto thinking system prompt to support it in train and inference, as shown in Figure~\ref{fig:sys_prompt}.
During the inference stage, controllable reasoning modes are achieved by prepending special tokens to the input, as depicted at the bottom of Figure \ref{fig:main}.

% 它是一个binary reward，正确匹配为1，否则 为0 （0/1）
\textbf{Answer Reward.}
We check the correctness of the tool call by comparing it against the ground-truth annotation $y^*$. 
Tool-call outputs are parsed into structured dictionaries, enabling exact matching of both the function name and all required arguments. 
% This involves checking whether the predicted tool name matches the ground-truth and whether all required arguments are present correctly. 
% This strict matching criterion ensures that the model learns to generate functionally precise and executable tool calls.
To encourage a balance between reasoning efficiency and accuracy, we design an asymmetric reward based on the mode (\textit{think} or \textit{no-think}):
\begin{equation}
    \small
    \mathcal{R}_{\text{answer}}(o_i) = 
    \begin{cases} 
    +1.0, & \text{if } o_i = y^*, \textit{no-think}, \\
    +0.5, & \text{if } o_i = y^*, \textit{think}, \\
    -0.5, & \text{if } o_i \neq y^*,  \textit{think}, \\
    -1.0, & \text{if } o_i \neq y^*,   \textit{no-think},
    \end{cases}
\end{equation}
It incentivizes short responses when they are correct, while encouraging long reasoning when mistakes occur, prompting more careful processing in uncertain scenarios.

% Therefore, the total reward for each completion is:
% \begin{equation}
% r_i = \mathcal{R}_{format}(o_i) + \mathcal{R}_{answer}(o_i),
% \end{equation}
% where both components are discretized, ensuring stable and interpretable credit assignment.
% % And $A_i$ is the advantage computed using a group of rewards $\{r_1, r_2, \dots, r_G\}$ corresponding to the completions within each group:
% The estimated advantage $A_i$  is then computed by normalizing the rewards within the group of $G$ rollouts:
% \begin{equation}
% A_i = \frac{r_i - \text{mean}(\{r_1, r_2, \dots, r_G\})}{\text{std}(\{r_1, r_2, \dots, r_G\})}.
% \end{equation}
% % This per-group advantage normalization reduces variance and enhances training stability by emphasizing relative performance among candidates.

\section{Experiments}
% In this section, we show the superiority of our method in performance and robustness across various benchmarks, and in-depth analysis to verify the effectiveness of our method.

\subsection{Experimental Setup}
\label{exp_setup}
We use the open-source Qwen2.5-7B-Instruct as our base model.
We compared four baseline types: \textbf{Base}, \textbf{SFT}-trained , API-based \textbf{Frontier}, and \textbf{RLVR}-trained models.
Additionally, we compare the series of models trained on the base model using \textit{PubTool} data.
See Appendix \ref{sec:appendix} for more details.

\noindent
\textbf{Evaluation Dataset}. 
The following benchmarks are used for evaluation: 
(1) \textbf{BFCL} \citep{bfclv3} provides a comprehensive dataset comprising 4k+ instances (updating), consisting of \textit{Non-live} (with expert-curated simple tools), \textit{Live} (with user-contributed complex tools),  \textit{Multi-turn} (with multi-turn \& multi-step tool use) samples.
(2) \textbf{API-Bank} \citep{li2023api}, which consists of 314 tool-use dialogues and 753 API calls. 
This dataset evaluates a models' abilities to correctly invoke a known API (L-1) based on a query and to retrieve and call APIs from a tool list (L-2).
% Here, \textit{Non-live} denotes simple tool use scenarios (e.g., single tool), while \textit{Live} represents more complex tool use scenarios (e.g., multiple parallel tools). 
(3) \textbf{ACEBench} \citep{chen2025acebench} is a 2k-entry benchmark for assessing agentic tool use, using its summary score in "normal" evaluation type (covering single-turn and multi-turn scenarios).
%else tau-bench, complextfuncall

\subsection{Overall Performance}
\begin{figure*}[th]
    \centering
  \includegraphics[width=0.8\linewidth]{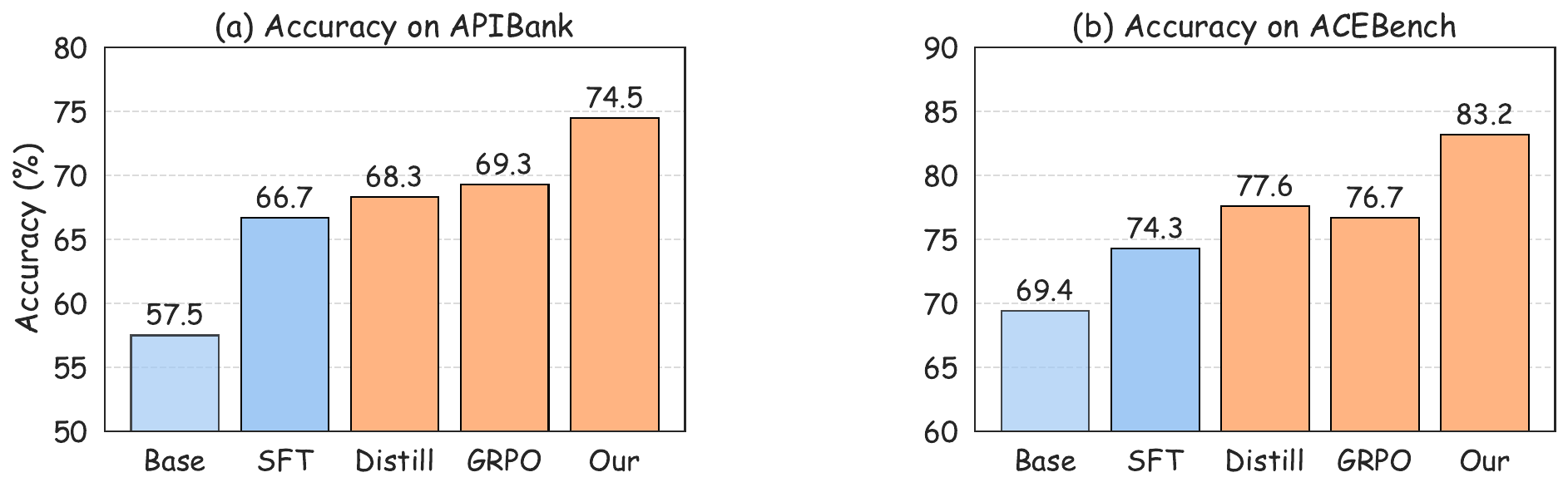}
  \caption{Performance of methods using training data \textit{PubTool} on APIBank and ACEBench.
  % \textcolor[HTML]{A9C8F9}{Bar} denotes short reasoning models, and \textcolor[HTML]{FDB483}{bar} denotes long reasoning models
    }
  \label{fig:exp3}
\end{figure*}

The overall performance of models are shown in Table \ref{overall_res} and Figure \ref{fig:exp3}.
Firstly, the results indicate that our model consistently achieves corresponding best performance at comparable scales (\textasciitilde7B).
For instance, compared to PubTool-SFT, AutoTool-7B with automatic think achieving +11.95 point improvement. 
And relative to Base model, it also has a remarkable boost with +16.43\%.
Secondly, our model demonstrated its more superiority in challenging scenarios (e.g., achieves +28.5\% improvement compare to PubTool-SFT in \textit{Multi-turn}).
This demonstrates that our method realizes a strong robustness enhancement in complex scenarios.

Moreover, our model outperforms most SFT-trained and RLVR-trained models in BFCL, and demonstrates comparable performance with the frontier models.
It also shows consistent advantageous performance on API-Bank and ACEBench compared with baselines in Figure~\ref{fig:exp3}. 
For example, on ACEBench, our model achieves a 6.5 improvement compared to GRPO and a 5.9 improvement compared to Distilled.
Finally, in the inference controllable mode, when forced to think, the overall performance is on par with auto think; 
when forced not to think, the effect on \textit{Multi-turn} is significantly improved compared to no-think models (e.g., PubTool-SFT).

\begin{table*}[th]
    \small
    \centering
    \caption{ Comparison on the BFCL benchmark.  \textit{Overall Acc} denotes the average performance on three subsets.
        \textsuperscript{*} indicates a single-turn tool use model;
        \textsuperscript{$\dagger$} denotes models trained on \textit{PubTool} data with a specific method.
        The subscript denotes the thinking rate.
        % \textbf{Bold} for best performance in R1-like models and \underline{underline} for best performance in the other types.
    }
    \begin{tabular}{l|l|ccc|c}
        \toprule
        Type & Model & Non-Live & Live  & Multi-Turn & \textbf{Overall Acc} \\  \midrule
        \multirow{3}{*}{\ding{168}Base }  
            & LLaMA-3.1-8B-Instruct & 84.21 & 61.08 & 9.62 & 50.87 \\
            & Qwen2.5-7B-Instruct & 86.46 & 67.44 & 7.62 & 53.69 \\ 
            & Qwen2.5-32B-Instruct & 85.81 & 74.23 & 17.75 & 59.67 \\
        \midrule
        \multirow{3}{*}{\ding{170}Frontier}  
            & GPT-4o-2024-11-20 & {87.67} & {79.88} & {43.00} & {70.42} \\
            & o3-2025-04-16 & 81.42 & 73.43 & 56.12 & 70.32 \\
            & Gemini-2.5-Pro & 89.54 & 76.83 & 30.62 & 65.48 \\
        
        \midrule
        \multirow{5}{*}{\ding{169}SFT}  
            & Hammer2.1-7b\citep{lin2024hammer} & {88.65} & 75.11 & 23.50 & 61.83 \\
            & ToolACE-8B \citep{liu2024toolace} & 87.54 & 78.59 & 7.75 & 58.42 \\
            & xLAM-7b-r\citep{zhang2024xlam} & 81.06 & 75.22 & 10.00 & 54.75 \\
            &  PubTool-SFT\textsuperscript{$\dagger$} & 88.98 & 77.28 & 9.68 & 58.17 \\
            & PubTool-Distilled\textsuperscript{$\dagger$} & 87.73 & 78.64 & 15.65 & 60.30 \\

        \midrule
        \multirow{6}{*}{\ding{171}RLVR }  & DeepSeek-R1-0528 & 75.20 & 77.30 & 38.88 & 63.79 \\
            & Qwen3-8B\citep{qwen3technicalreport} & 88.81 & 78.54 & 33 & 66.34 \\
            % & Qwen3-32B & 88.90 & 77.83 & 43.12 & 69.25 \\
            &  QwQ-32B\citep{qwq32b} & 87.33 & 75.61 & 14.50 & 58.30 \\ 
            &  Tool-N1-7B\textsuperscript{*}\citep{zhang2025nemotron} & 89.25 & 80.38 &  -  & - \\ 
            % &  Tool-N1-14B\textsuperscript{*} & 90.52 & 81.42 &  -  & - \\ 
            & ToolRL-7B\citep{qian2025toolrl} & 82.21 & 74.90 & 18.12  & 58.38 \\ 
            & Thinkless\citep{fang2025thinkless} & 86.92 & 77.62 &	24.64 &	63.06 \\ 
            & Adactrl\citep{huang2025adactrl} & 86.36	& 73.12	& 15.63	& 58.37 \\ 
            & PubTool-GRPO\textsuperscript{$\dagger$} & 88.87 & 78.93 & 10.77 & 60.13 \\
        
        \midrule
        \multirow{3}{*}{\ding{171}Ours} 
        & {AutoTool-7B\textsuperscript{$\dagger$}} & {89.76\textsubscript{0\%}} & {80.22\textsubscript{4.8\%}} & {38.18\textsubscript{45\%}} & {70.12\textsubscript{9.7\%}} \\
        & \hspace{0.2cm} \textit{+ think } & 89.86 & 80.43 & 39.28 & 70.71 \\  
        & \hspace{0.2cm} \textit{+ no-think } & 87.36 & 78.60 & 27.63 & 63.34 \\  
        \bottomrule
        % \multicolumn{3}{l}{\(\clubsuit\)\textit{Reasoning models}} & \\ \midrule  \rowcolor{blue!20} 
        % \rowcolor{green!20} 1 & 62.92 & \(\clubsuit\) \textbf{AutoTool-7B} & 88.82 & 76.82 & 23.84 & 84.90/80.72 \\
    \end{tabular}
    \label{overall_res}
\end{table*}

% 分析实验
% 1. 计算效率，思考率。分析
    % 自动思考 和 不思考 ，全思考的效果 对比，在主实验表格中 给出即可。
% 2。 消融实验：预热分析，数据过滤问题，无dynamic熵loss，all 不解耦，
    % 原始 dapo + our dapo //不同的算法 dapo 试下效果。
% 3. Visualization 训练过程 
% 4. //不同模型规模的分析，2.5 1.5-32B，qwen3 7B?

\subsection{Deep  Analysis Study}
\subsubsection{Ablation Study}
To evaluate the effectiveness of key components in our method, we conducted an ablation study with the following variations:
(1) Replaced the adaptive entropy coefficient with a fixed one (\textit{w/o adapt coeff});
(2) Replaced the decoupling loss with a unified loss with fixed entropy constraint (\textit{w/o decouple});
(3) Removed data quality refinement (\textit{w/o data refine}).
We also included Qwen2.5-7B-Instruct as a {Base Model} for comparison. 
As shown in Table \ref{tab:ablation},
compared with the baseline, our full model delivers a significant improvement of 16.43 points in Overall performance.
All components are essential to our method, and removing any component causes clear performance drops: 
(1) \textit{w/o data refine} brings the largest 6.43\% Overall reduction, highlighting high-quality data as a core foundation. 
(2) \textit{w/o adapt coeff} leads to a 10.53\% Multi-turn decline, proving its value in stabilizing multi-round interactions; 
(3) \textit{w/o decouple} results in a 2.34\% Overall drop, showing decoupling avoids objective interference.

\begin{table}[th]
    \centering
    % \small
    \caption{The strategy ablation performance (↑ = increase, ↓ = decrease, values are relative percentage changes from the \textit{Our (w/. all)} model). }
    \label{tab:ablation}
    \begin{tabular}{l|ccc|c}
        \toprule
        Models & Non-live & Live & Multi-turn & Overall\\ 
        \midrule
        Base Model & 86.46 & 67.44 & 7.62 & 53.69 \\ 
        \midrule
        Our (w/. all)  & 89.76 & 80.22 & 38.18 & 70.12 \\
        \hspace{0.2cm}w/o. \textit{data refine} & 88.22 \textcolor{red!70!black}{\textsubscript{↓1.54}} & 73.29 \textcolor{red!70!black}{\textsubscript{↓6.93}} & 26.84 \textcolor{red!70!black}{\textsubscript{↓11.34}} & 63.69 \textcolor{red!70!black}{\textsubscript{↓6.43}} \\ 
        \hspace{0.2cm}w/o. \textit{decouple} & 87.35 \textcolor{red!70!black}{\textsubscript{↓2.41}} & 75.98 \textcolor{red!70!black}{\textsubscript{↓4.24}} & 27.65 \textcolor{red!70!black}{\textsubscript{↓10.53}} & 64.23 \textcolor{red!70!black}{\textsubscript{↓5.89}} \\
        \hspace{0.2cm}w/o. \textit{adapt coeff} & 88.73 \textcolor{red!70!black}{\textsubscript{↓1.03}} & 78.73 \textcolor{red!70!black}{\textsubscript{↓1.49}} & 32.14 \textcolor{red!70!black}{\textsubscript{↓6.04}} & 67.78 \textcolor{red!70!black}{\textsubscript{↓2.34}} \\
        \bottomrule
    \end{tabular}
\end{table}

% 为了分析模型的推理效率，
\subsubsection{Inference Efficiency analysis}
Given the trade-off between reasoning path length, model size (\textasciitilde B), and performance, we introduce a new metric, Accuracy per Computation Unit (ACU), to better capture this balance and assess model inference efficiency \citep{ma2025cot}.
It is defined as:
\begin{equation}
\small
\text{ACU} = \frac{\text{Accuracy}}{\#\text{Params} \times \#\text{Tokens}}
\end{equation}
Since the ACU value typically falls within the range of \(10^{-5}\) to \(10^{-3}\), we report it in units of \(10^3\) for improved readability.
In addition, we report the thinking rates of our model across all submetrics.

% 转成pdf后，划一个线在a,b 之间。
\begin{figure*}[t]
    \centering
    \includegraphics[width=0.85\linewidth]{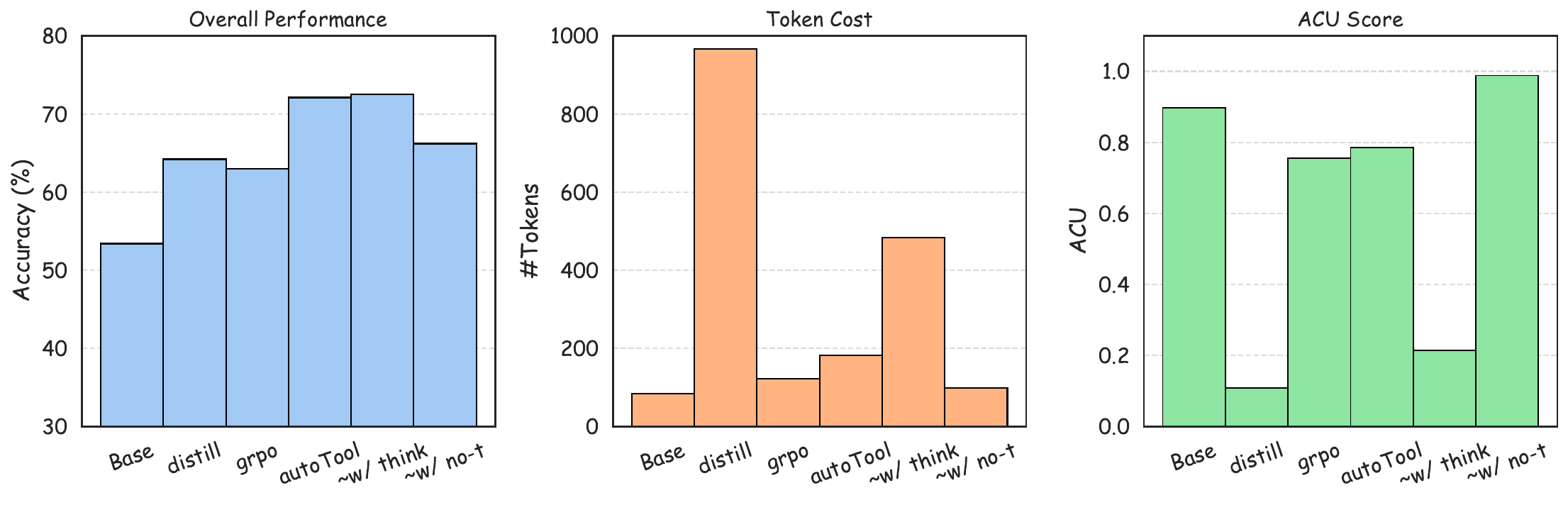}
    % \begin{subfigure}[t]{0.7\linewidth}
    %     \centering
    %     \includegraphics[width=1.0\linewidth]{./exp1.pdf}
    %     \caption{Inference Efficiency}
    %     \label{fig:exp2a}
    % \end{subfigure}
    % % \hfill
    % \begin{subfigure}[t]{0.29\linewidth}
    %     \centering
    %     \includegraphics[width=0.95\linewidth]{./exp2.pdf}
    %     \caption{Thinking Rate}
    %     \label{fig:exp2b}
    % \end{subfigure}
  \caption{
      Inference efficiency analysis results, including performance, token cost, ACU.}
  \label{tab:efficiency}
\end{figure*}

The experimental results are summarized in Figure \ref{tab:efficiency}. 
From the results, we observe that AutoTool achieves the second-best overall performance: it reduces token cost significantly by 81\%, requiring only about \textasciitilde183 tokens compared to the distilled model (\textasciitilde966 tokens). 
Notably, with the forced no-think inference mode, AutoTool attains the optimal ACU score (0.97), demonstrating excellent inference efficiency. 
Even with the think inference mode, it still delivers the highest accuracy while cutting the token cost by half relative to the distilled model. 
% This suggests that it focuses on enhancing performance in complex tool-use cases, potentially at the expense of simpler tool-use cases., therefore, 有高的推理效率。
Additionally, Table \ref{overall_res} shows that the our model's thinking rate reaches 45\% in the Multi-Turn scenario but 0\% in the No-Live scenario. 
The training process visualized in Appendix Section~\ref{sec:visualization} shows our model extends reasoning trajectories for complex questions by $\times 5$, while enabling concise responses for simple ones.
This suggests the model has learned to automatically adjust the test-time scale based on sample difficulty, which effectively supports the improvement of inference efficiency.

\section{Conclusion}
This study focused on addressing challenges in integrating agentic LLMs with tools by optimizing the RLVR paradigm.
Our research first identified two critical issues: excessive resource consumption caused by unnecessary long-trajectory reasoning, and the \textit{reasoning collapse} phenomenon under the direct RL training, hindering effective scaling.
To solve these, we proposed a decoupled adaptive entropy constraint strategy, which enables the model to automatically adjust reasoning scales based on problem difficulty, thereby balancing performance and inference efficiency.
Experiments on three benchmarks confirmed the strategy’s effectiveness, boosting accuracy while cutting inference token cost significantly.
This work advances RL-based agentic tool-use training and provides a practical auto-scaling solution for efficiently handling tasks.

% \section*{Limitation}
% While our study has achieved notable advancements, it is important to acknowledge several limitations that could be addressed in future work.
% (1) The progressive reward-switching strategy, though effective for generalization, introduces additional computational costs during the RL training phase, particularly for large models (e.g., Tool-Zero-32B). This limits scalability on resource-constrained hardware without further optimization.
% (2) We acknowledge that evaluation datasets (e.g., BFCL and API-Bank) have known limitations (e.g., lengthy calling chains) arising from design preferences.
% We will address these limitations in our future work.

\section*{Acknowledgements}
The research in this article is supported by the New Generation Artificial Intelligence of China (2024YFE0203700), National Natural Science Foundation of China under Grants U22B2059 and 62576124.

\section*{Ethics Statement}
This work strictly adheres to the ICLR Code of Ethics: it involves no human subjects, uses datasets compliant with original licensing agreements (ensuring privacy and legal compliance), and avoids discriminatory biases in experimental design/results; all authors confirm adherence, with no conflicting sponsorships. 

\section*{Reproducibility Statement}
For reproducibility, key details are referenced across the main text (methodology Section~\ref{sec:method}, experimental setup Section~\ref{exp_setup}), appendix (hyperparameters Section~\ref{sec:appendix}, data processing details Section~\ref{sec:dataset}, full prompt Section~\ref{sec:auto_prompt}), and supplementary materials (anonymous source code). 
We ensure data splits, random seeds, and environment configurations are explicitly stated, allowing researchers to independently verify our findings under identical conditions.

% Use unnumbered third level headings for the acknowledgments. 
% All acknowledgments, including those to funding agencies, go at the end of the paper.
\bibliography{iclr2026_conference,custom}

@misc{deepseekai2025r1,
      title={DeepSeek-R1: Incentivizing Reasoning Capability in LLMs via Reinforcement Learning}, 
      author={DeepSeek-AI},
      year={2025},
      eprint={2501.12948},
      archivePrefix={arXiv},
      primaryClass={cs.CL},
      url={https://arxiv.org/abs/2501.12948}, 
}

@inproceedings{bfclv3,
    title={Berkeley Function Calling Leaderboard},
    author={Fanjia Yan and Huanzhi Mao and Charlie Cheng-Jie Ji and Tianjun Zhang and Shishir G. Patil and Ion Stoica and Joseph E. Gonzalez},
    year={2024},
    howpublished={\url{https://gorilla.cs.berkeley.edu/blogs/8_berkeley_function_calling_leaderboard.html}},
    }

@inproceedings{toolllm,
  title={ToolLLM: Facilitating Large Language Models to Master 16000+ Real-world APIs},
  author={Qin, Yujia and Liang, Shihao and Ye, Yining and Zhu, Kunlun and Yan, Lan and Lu, Yaxi and Lin, Yankai and Cong, Xin and Tang, Xiangru and Qian, Bill and others},
  year = {2023},
  booktitle={The Twelfth International Conference on Learning Representations}
}

@article{qu2025tool,
  title={Tool learning with large language models: A survey},
  author={Qu, Changle and Dai, Sunhao and Wei, Xiaochi and Cai, Hengyi and Wang, Shuaiqiang and Yin, Dawei and Xu, Jun and Wen, Ji-Rong},
  journal={Frontiers of Computer Science},
  volume={19},
  number={8},
  pages={198343},
  year={2025},
  publisher={Springer}
}

@inproceedings{wang2024executable,
  title={Executable code actions elicit better llm agents},
  author={Wang, Xingyao and Chen, Yangyi and Yuan, Lifan and Zhang, Yizhe and Li, Yunzhu and Peng, Hao and Ji, Heng},
  booktitle={Forty-first International Conference on Machine Learning},
  year={2024}
}

@article{lazaridou2022internet,
  title={Internet-augmented language models through few-shot prompting for open-domain question answering},
  author={Lazaridou, Angeliki and Gribovskaya, Elena and Stokowiec, Wojciech and Grigorev, Nikolai},
  journal={arXiv preprint arXiv:2203.05115},
  year={2022}
}

@article{shuster2022blenderbot,
  title={Blenderbot 3: a deployed conversational agent that continually learns to responsibly engage},
  author={Shuster, Kurt and Xu, Jing and Komeili, Mojtaba and Ju, Da and Smith, Eric Michael and Roller, Stephen and Ung, Megan and Chen, Moya and Arora, Kushal and Lane, Joshua and others},
  journal={arXiv preprint arXiv:2208.03188},
  year={2022}
}

@article{nakano2021webgpt,
  title={Webgpt: Browser-assisted question-answering with human feedback},
  author={Nakano, Reiichiro and Hilton, Jacob and Balaji, Suchir and Wu, Jeff and Ouyang, Long and Kim, Christina and Hesse, Christopher and Jain, Shantanu and Kosaraju, Vineet and Saunders, William and others},
  journal={arXiv preprint arXiv:2112.09332},
  year={2021}
}

@article{song2024adaptive,
  title={Adaptive in-conversation team building for language model agents},
  author={Song, Linxin and Liu, Jiale and Zhang, Jieyu and Zhang, Shaokun and Luo, Ao and Wang, Shijian and Wu, Qingyun and Wang, Chi},
  journal={arXiv preprint arXiv:2405.19425},
  year={2024}
}

@article{chen2022program,
  title={Program of thoughts prompting: Disentangling computation from reasoning for numerical reasoning tasks},
  author={Chen, Wenhu and Ma, Xueguang and Wang, Xinyi and Cohen, William W},
  journal={arXiv preprint arXiv:2211.12588},
  year={2022}
}

@article{liu2024toolace,
  title={Toolace: Winning the points of llm function calling},
  author={Liu, Weiwen and Huang, Xu and Zeng, Xingshan and Hao, Xinlong and Yu, Shuai and Li, Dexun and Wang, Shuai and Gan, Weinan and Liu, Zhengying and Yu, Yuanqing and others},
  journal={arXiv preprint arXiv:2409.00920},
  year={2024}
}

@article{prabhakar2025apigen,
  title={APIGen-MT: Agentic PIpeline for Multi-Turn Data Generation via Simulated Agent-Human Interplay},
  author={Prabhakar, Akshara and Liu, Zuxin and Zhu, Ming and Zhang, Jianguo and Awalgaonkar, Tulika and Wang, Shiyu and Liu, Zhiwei and Chen, Haolin and Hoang, Thai and others},
  journal={arXiv preprint arXiv:2504.03601},
  year={2025}
}

@article{zhang2024ecoact,
  title={EcoAct: Economic Agent Determines When to Register What Action},
  author={Zhang, Shaokun and Zhang, Jieyu and Ding, Dujian and Garcia, Mirian Hipolito and Mallick, Ankur and Madrigal, Daniel and Xia, Menglin and R{\"u}hle, Victor and Wu, Qingyun and Wang, Chi},
  journal={arXiv preprint arXiv:2411.01643},
  year={2024}
}

@article{zhang2024xlam,
  title={xLAM: A Family of Large Action Models to Empower AI Agent Systems}, 
  author={Zhang, Jianguo  and Lan, Tian  and Zhu, Ming  and Liu, Zuxin and Hoang, Thai and Kokane, Shirley and Yao, Weiran and Tan, Juntao and Prabhakar, Akshara and Chen, Haolin and Liu, Zhiwei and Feng, Yihao and Awalgaonkar, Tulika and Murthy, Rithesh and Hu, Eric and Chen, Zeyuan and Xu, Ran and Niebles, Juan Carlos and Heinecke, Shelby and Wang, Huan and Savarese, Silvio and Xiong, Caiming},
  journal={arXiv preprint arXiv:2409.03215},
  year={2024}
}

@article{zeng2025boosting,
  title={Boosting Tool Use of Large Language Models via Iterative Reinforced Fine-Tuning},
  author={Zeng, Yirong and Ding, Xiao and Wang, Yuxian and Liu, Weiwen and Ning, Wu and Hou, Yutai and Huang, Xu and Qin, Bing and Liu, Ting},
  journal={arXiv preprint arXiv:2501.09766},
  year={2025}
}

@article{yu2024steptool,
  title={StepTool: A Step-grained Reinforcement Learning Framework for Tool Learning in LLMs},
  author={Yu, Yuanqing and Wang, Zhefan and Ma, Weizhi and Guo, Zhicheng and Zhan, Jingtao and Wang, Shuai and Wu, Chuhan and Guo, Zhiqiang and Zhang, Min},
  journal={arXiv preprint arXiv:2410.07745},
  year={2024}
}

@article{muennighoff2025s1,
  title={s1: Simple test-time scaling},
  author={Muennighoff, Niklas and Yang, Zitong and Shi, Weijia and Li, Xiang Lisa and Fei-Fei, Li and Hajishirzi, Hannaneh and Zettlemoyer, Luke and Liang, Percy and Cand{\`e}s, Emmanuel and Hashimoto, Tatsunori},
  journal={arXiv preprint arXiv:2501.19393},
  year={2025}
}

@article{pan2025metaspatial,
  title={MetaSpatial: Reinforcing 3D Spatial Reasoning in VLMs for the Metaverse},
  author={Pan, Zhenyu and Liu, Han},
  journal={arXiv preprint arXiv:2503.18470},
  year={2025}
}

@article{xia2025generative,
  title={Generative AI Act II: Test Time Scaling Drives Cognition Engineering},
  author={Xia, Shijie and Qin, Yiwei and Li, Xuefeng and Ma, Yan and Fan, Run-Ze and Chern, Steffi and Zou, Haoyang and Zhou, Fan and Hu, Xiangkun and Jin, Jiahe and others},
  journal={arXiv preprint arXiv:2504.13828},
  year={2025}
}

@article{shao2024deepseekmath,
  title={Deepseekmath: Pushing the limits of mathematical reasoning in open language models},
  author={Shao, Zhihong and Wang, Peiyi and Zhu, Qihao and Xu, Runxin and Song, Junxiao and Bi, Xiao and Zhang, Haowei and Zhang, Mingchuan and Li, YK and Wu, Y and others},
  journal={arXiv preprint arXiv:2402.03300},
  year={2024}
}

@article{yu2025dapo,
  title={Dapo: An open-source llm reinforcement learning system at scale},
  author={Yu, Qiying and Zhang, Zheng and Zhu, Ruofei and Yuan, Yufeng and Zuo, Xiaochen and Yue, Yu and Fan, Tiantian and Liu, Gaohong and Liu, Lingjun and Liu, Xin and others},
  journal={arXiv preprint arXiv:2503.14476},
  year={2025}
}

@article{li2025torl,
  title={Torl: Scaling tool-integrated rl},
  author={Li, Xuefeng and Zou, Haoyang and Liu, Pengfei},
  journal={arXiv preprint arXiv:2503.23383},
  year={2025}
}

@article{feng2025retool,
  title={Retool: Reinforcement learning for strategic tool use in llms},
  author={Feng, Jiazhan and Huang, Shijue and Qu, Xingwei and Zhang, Ge and Qin, Yujia and Zhong, Baoquan and Jiang, Chengquan and Chi, Jinxin and Zhong, Wanjun},
  journal={arXiv preprint arXiv:2504.11536},
  year={2025}
}

@article{jin2025search,
  title={Search-r1: Training llms to reason and leverage search engines with reinforcement learning},
  author={Jin, Bowen and Zeng, Hansi and Yue, Zhenrui and Yoon, Jinsung and Arik, Sercan and Wang, Dong and Zamani, Hamed and Han, Jiawei},
  journal={arXiv preprint arXiv:2503.09516},
  year={2025}
}

@article{meng2025mm,
  title={Mm-eureka: Exploring visual aha moment with rule-based large-scale reinforcement learning},
  author={Meng, Fanqing and Du, Lingxiao and Liu, Zongkai and Zhou, Zhixiang and Lu, Quanfeng and Fu, Daocheng and Shi, Botian and Wang, Wenhai and He, Junjun and Zhang, Kaipeng and others},
  journal={CoRR},
  year={2025}
}

@misc{ds0528,
      title={DeepSeek-R1: Incentivizing Reasoning Capability in LLMs via Reinforcement Learning}, 
      author={DeepSeek-AI},
      year={2025},
      eprint={2501.12948},
      archivePrefix={arXiv},
      primaryClass={cs.CL},
      url={https://arxiv.org/abs/2501.12948}, 
}

@misc{openr1,
    title = {Open R1: A fully open reproduction of DeepSeek-R1},
    url = {https://github.com/huggingface/open-r1},
    author = {{Hugging Face}},
    month = {January},
    year = {2025}
}

@article{he2025skywork,
  title={Skywork Open Reasoner 1 Technical Report},
  author={He, Jujie and Liu, Jiacai and Liu, Chris Yuhao and Yan, Rui and Wang, Chaojie and Cheng, Peng and Zhang, Xiaoyu and Zhang, Fuxiang and Xu, Jiacheng and Shen, Wei and Li, Siyuan and Zeng, Liang and Wei, Tianwen and Cheng, Cheng and An, Bo and Liu, Yang and Zhou, Yahui},
  journal={arXiv preprint arXiv:2505.22312},
  year={2025}
}

@misc{Hermes-Function-Calling-Dataset-V1, 
      url={https://huggingface.co/NousResearch/hermes-function-calling-v1},
      title={Hermes-Function-Calling-Dataset-V1}, 
      author={interstellarninja, Teknium},
      year = {2024}
}

@article{li2025limr,
  title={Limr: Less is more for rl scaling},
  author={Li, Xuefeng and Zou, Haoyang and Liu, Pengfei},
  journal={arXiv preprint arXiv:2502.11886},
  year={2025}
}

@inproceedings{li2023api,
  title={API-Bank: A Comprehensive Benchmark for Tool-Augmented LLMs},
  author={Li, Minghao and Zhao, Yingxiu and Yu, Bowen and Song, Feifan and Li, Hangyu and Yu, Haiyang and Li, Zhoujun and Huang, Fei and Li, Yongbin},
  booktitle={Proceedings of the 2023 Conference on Empirical Methods in Natural Language Processing},
  pages={3102--3116},
  year={2023}
}

@article{lin2024hammer,
  title={Hammer: Robust function-calling for on-device language models via function masking},
  author={Lin, Qiqiang and Wen, Muning and Peng, Qiuying and Nie, Guanyu and Liao, Junwei and Wang, Jun and Mo, Xiaoyun and Zhou, Jiamu and Cheng, Cheng and Zhao, Yin and others},
  journal={arXiv preprint arXiv:2410.04587},
  year={2024}
}

@misc{qwq32b,
    title = {QwQ-32B: Embracing the Power of Reinforcement Learning},
    url = {https://qwenlm.github.io/blog/qwq-32b/},
    author = {Qwen Team},
    month = {March},
    year = {2025}
}

@misc{qwen3technicalreport,
      title={Qwen3 Technical Report}, 
      author={Qwen Team},
      year={2025},
      eprint={2505.09388},
      archivePrefix={arXiv},
      primaryClass={cs.CL},
      url={https://arxiv.org/abs/2505.09388}, 
}

@article{chen2025acebench,
  title={ACEBench: Who Wins the Match Point in Tool Learning?},
  author={Chen, Chen and Hao, Xinlong and Liu, Weiwen and Huang, Xu and Zeng, Xingshan and Yu, Shuai and Li, Dexun and Wang, Shuai and Gan, Weinan and Huang, Yuefeng and others},
  journal={arXiv e-prints},
  pages={arXiv--2501},
  year={2025}
}

@article{qian2025toolrl,
  title={Toolrl: Reward is all tool learning needs},
  author={Qian, Cheng and Acikgoz, Emre Can and He, Qi and Wang, Hongru and Chen, Xiusi and Hakkani-T{\"u}r, Dilek and Tur, Gokhan and Ji, Heng},
  journal={arXiv preprint arXiv:2504.13958},
  year={2025}
}

@article{zhang2025nemotron,
  title={Nemotron-Research-Tool-N1: Tool-Using Language Models with Reinforced Reasoning},
  author={Zhang, Shaokun and Dong, Yi and Zhang, Jieyu and Kautz, Jan and Catanzaro, Bryan and Tao, Andrew and Wu, Qingyun and Yu, Zhiding and Liu, Guilin},
  journal={arXiv preprint arXiv:2505.00024},
  year={2025}
}

@article{fang2025thinkless,
  title={Thinkless: Llm learns when to think},
  author={Fang, Gongfan and Ma, Xinyin and Wang, Xinchao},
  journal={arXiv preprint arXiv:2505.13379},
  year={2025}
}

@article{zhang2025adaptthink,
  title={Adaptthink: Reasoning models can learn when to think},
  author={Zhang, Jiajie and Lin, Nianyi and Hou, Lei and Feng, Ling and Li, Juanzi},
  journal={arXiv preprint arXiv:2505.13417},
  year={2025}
}

@article{huang2025adactrl,
  title={AdaCtrl: Towards Adaptive and Controllable Reasoning via Difficulty-Aware Budgeting},
  author={Huang, Shijue and Wang, Hongru and Zhong, Wanjun and Su, Zhaochen and Feng, Jiazhan and Cao, Bowen and Fung, Yi R},
  journal={arXiv preprint arXiv:2505.18822},
  year={2025}
}

@article{wang2025think,
  title={Think or Not? Selective Reasoning via Reinforcement Learning for Vision-Language Models},
  author={Wang, Jiaqi and Lin, Kevin Qinghong and Cheng, James and Shou, Mike Zheng},
  journal={arXiv preprint arXiv:2505.16854},
  year={2025}
}

@article{ma2025cot,
  title={Cot-valve: Length-compressible chain-of-thought tuning},
  author={Ma, Xinyin and Wan, Guangnian and Yu, Runpeng and Fang, Gongfan and Wang, Xinchao},
  journal={arXiv preprint arXiv:2502.09601},
  year={2025}
}

@misc{ragen,
      title={RAGEN: Understanding Self-Evolution in LLM Agents via Multi-Turn Reinforcement Learning}, 
      author={Zihan, Wang and Kangrui, Wang and Qineng, Wang and Pingyue, Zhang and Linjie, Li and Zhengyuan, Yang and Xing, Jin and Kefan, Yu and Minh Nhat Nguyen and Licheng, Liu and Eli Gottlieb and Yiping Lu and Kyunghyun, Cho and Jiajun, Wu and Li, Fei-Fei and Lijuan, Wang and Yejin,Choi and Manling, Li},
      year={2025},
      eprint={2504.20073},
      archivePrefix={arXiv},
      primaryClass={cs.LG},
      url={https://arxiv.org/abs/2504.20073}, 
}

@misc{Agent-R1,
  author       = {Jie Ouyang and Ruiran Yan and Yucong Luo and Mingyue Cheng and Qi Liu and Zirui Liu and Shuo Yu and Daoyu Wang},
  title        = {Training Powerful LLM Agents with End-to-End Reinforcement Learning},
  year         = {2025},
  organization = {GitHub},
  url          = {https://github.com/0russwest0/Agent-R1},
}

@misc{KimiResearcher2024,
  author       = {Kimi K2},
  title        = {{Kimi-researcher}: End-to-end RL Training for Emerging Agentic Capabilities},
  year         = {2024},
  howpublished = {\url{https://moonshotai.github.io/Kimi-Researcher/}},
  note         = {Accessed: 2025-09-10},
}

@misc{OpenAIDeepResearch2025,
  author       = {{OpenAI}},
  title        = {o3},
  year         = {2025},
  howpublished = {\url{https://openai.com/index/introducing-deep-research/}},
  note         = {Accessed: Sep 10, 2025}
}

@misc{zeng2025simplerl2,
      title={7B Model and 8K Examples: Emerging Reasoning with Reinforcement Learning is Both Effective and Efficient},
      author={Weihao Zeng and Yuzhen Huang and Wei Liu and Keqing He and Qian Liu and Zejun Ma and Junxian He},
      year={2025},
      howpublished={\url{https://hkust-nlp.notion.site/simplerl-reason}},
      note={Notion Blog}
}

@article{zeng2025simplerl,
  title={Simplerl-zoo: Investigating and taming zero reinforcement learning for open base models in the wild},
  author={Zeng, Weihao and Huang, Yuzhen and Liu, Qian and Liu, Wei and He, Keqing and Ma, Zejun and He, Junxian},
  journal={arXiv preprint arXiv:2503.18892},
  year={2025}
}

@article{schulman2017proximal,
  title={Proximal policy optimization algorithms},
  author={Schulman, John and Wolski, Filip and Dhariwal, Prafulla and Radford, Alec and Klimov, Oleg},
  journal={arXiv preprint arXiv:1707.06347},
  year={2017}
}

@article{engstrom2020implementation,
  title={Implementation matters in deep policy gradients: A case study on ppo and trpo},
  author={Engstrom, Logan and Ilyas, Andrew and Santurkar, Shibani and Tsipras, Dimitris and Janoos, Firdaus and Rudolph, Larry and Madry, Aleksander},
  journal={arXiv preprint arXiv:2005.12729},
  year={2020}
}
\bibliographystyle{iclr2026_conference}

\appendix
% \section{Appendix}
% You may include other additional sections here.
\section*{Use of LLM}
LLMs (GPT-4o) were only used as general-purpose tools to draft baseline literature summaries and proofread minor grammar, no LLM contributed to core ideation, algorithm development, analysis, or writing, and all LLM-assisted content was verified for accuracy/integrity. No LLM is eligible for authorship.

\section{Related Work}
\label{sec:relatedwork}
\subsection{Agentic Tool-Use}
Enhancing LLMs with external tools has emerged as a pivotal direction for addressing complex tasks in open domains \citep{qu2025tool, wang2024executable}. Typical applications include integrating LLMs with search engines \citep{zhang2024ecoact, lazaridou2022internet, shuster2022blenderbot}, calculators \citep{nakano2021webgpt}, and Python interpreters \citep{wang2024executable, song2024adaptive, chen2022program}. 
Three common paradigms are widely adopted for training tool-use LLMs:
(1) SFT: imitates the reasoning patterns from labeled high-quality examples, enabling models to learn standard tool-use workflows \citep{liu2024toolace,zhang2024xlam,toolllm,prabhakar2025apigen}.
(2) RL with direct preference optimization: aligns model tool-use behavior with human intentions by optimizing against human preference signals \citep{zeng2025boosting, yu2024steptool}.
(3) RL with Verifiable Rewards (RLVR): as a novel approach, leverages scalable test-time inference and utilizes verifiable signals as rewards to refine the model's tool-use decisions \citep{li2025torl}.

\subsection{RL Scale-up}
Reinforcement learning (RL) has gained traction as a more scalable and generalizable training paradigm. 
Models like R1-Zero leverage group relative policy optimization (GRPO) \citep{shao2024deepseekmath} to unlock the model’s reasoning capabilities at test time \citep{deepseekai2025r1, yu2025dapo}. 
This R1-style reasoning paradigm, marking a shift from train-time scaling to test-time scaling \citep{muennighoff2025s1, xia2025generative}, has demonstrated success in mathematics \citep{shao2024deepseekmath}, coding \citep{pan2025metaspatial}, and agentic tool use \citep{feng2025retool, jin2025search}.

Recently, several works have explored automatic scaling , i.e., enabling models to adaptively select the optimal reasoning mode based on problem difficulty \citep{fang2025thinkless,zhang2025adaptthink,huang2025adactrl,wang2025think}.
In agentic tool-use tasks, auto-scaling is particularly critical: many such problems can be solved with short reasoning, whereas excessively long reasoning leads to unnecessary resource consumption.
While RL-based scaling for tool use in open-domain reasoning has been investigated \citep{zhang2025nemotron,qian2025toolrl}, RL with auto-scaling remains unexplored in agentic tool use.
% Nemotron-Research-Tool-N1: Tool-Using Language Models with Reinforced Reasoning
% ToolRL: Reward is All Tool Learning Needs
% Agent RL Scaling Law: Spontaneous Code Execution for Mathematical Problem Solving 代码块执行==》数学
% SkyRL-v0: Train Real-World Long-Horizon Agents via Reinforcement Learning
% RAGEN: Understanding Self-Evolution in LLM Agents via Multi-Turn Reinforcement Learning

\section{Details in Data Preparation }
\label{sec:dataset}

\textbf{Source of Training Data Details}. 
The raw data was sourced as follows:
\begin{itemize}
    \item ToolACE \citep{liu2024toolace}: A general tool-use dataset teaching models when to invoke tools vs. respond directly, enhancing multi-step decision-making.
    \item xLAM \citep{zhang2024xlam,prabhakar2025apigen}: A compositional dataset requiring one or more tool calls per turn. We mixed the original 60k xLAM with its multi-turn variant APIGen-MT-5k \citep{prabhakar2025apigen}.
    \item Hermes Function-Calling \citep{Hermes-Function-Calling-Dataset-V1}: Designed to train LLMs in function calls and structured output from natural language. We extracted function call-related samples.
\end{itemize}
The dataset features various conversational scenarios where AI agents are required to interpret queries and execute appropriate single or multiple function calls. 
In Section \ref{sec:pre_study}, data distillation employs Deepseek-R1-0528 \citep{deepseekai2025r1}. 
Subsequently, in Section \S \ref{sec:dataprepare}, to mitigate model bias by aligning with a no-think model, data distillation is carried out using Qwen3-32B \citep{qwen3technicalreport}.

\begin{figure}[th]
    \centering
  \begin{subfigure}[t]{0.45\linewidth}
        \centering
        \includegraphics[width=0.8\linewidth]{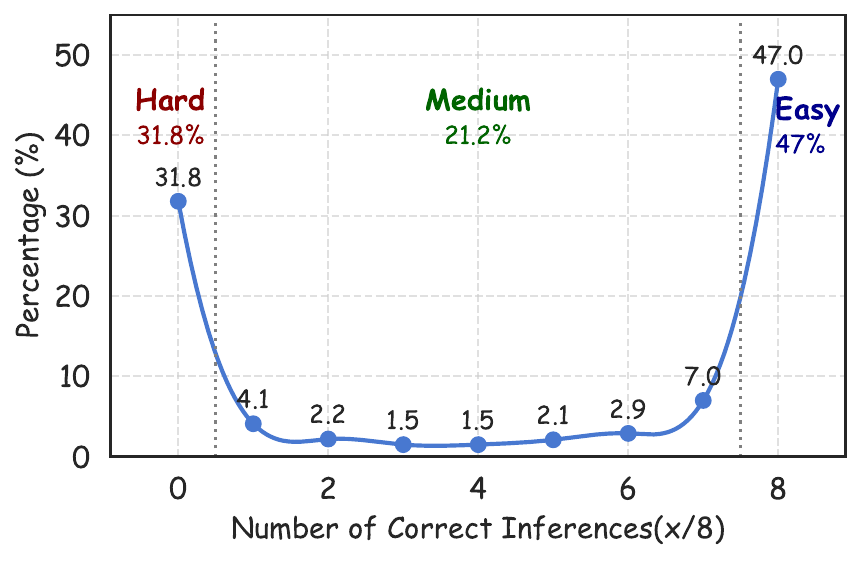}
        \caption{Data Distribution of Raw data}
        \label{fig:appx1a}
    \end{subfigure}
    % \hfill
    \begin{subfigure}[t]{0.45\linewidth}
        \centering
        \includegraphics[width=0.8\linewidth]{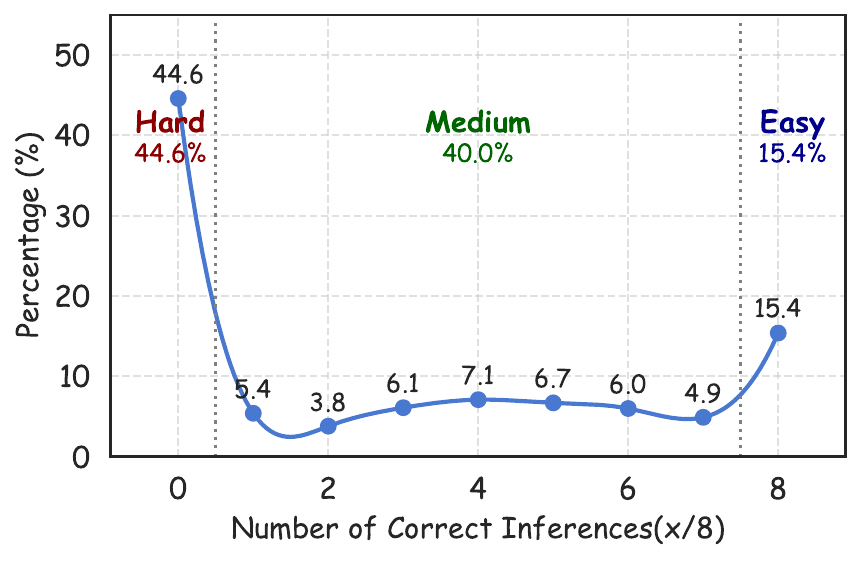}
        \caption{Data Distribution of PubTool}
        \label{fig:appx1b}
    \end{subfigure}
  \caption{The number of correct inferences distribution with performing 8 rounds of reasoning on the raw training data (a). The distribution of PubTool after data processing (b).
}
  \label{fig:data_dist}
\end{figure}

\textbf{Data Processing Pipeline \& Distribution Details.}
We obtained PubTool from raw data through following data processing workflow:
(1) We randomly downsampled xLAM to balance the sample sizes across the three datasets.
(2) We removed overly simple and excessively difficult samples;
Figure~\ref{fig:data_dist} shows the raw-data distribution of successful reasoning counts (pass@8).
Guided by this distribution, we partitioned the data into hard (31.8\%), medium (21.2\%), and easy (47\%) subsets (Figure~\ref{fig:data_dist}a) and re-balanced the difficulty distribution accordingly.
(3) For RL data, we further refined the set by prioritizing samples that align closely with the model’s current learning trajectory (details in the next paragraph).
For comparison, we visualized the PubTool distribution in the same way (Figure~\ref{fig:data_dist}b).
We observed that the original corpus is concentrated in the easy and hard extremes, whereas PubTool peaks in the hard subset and is sparse in the easy subset.
We argue that training on data of moderately high difficulty better elicits the model’s test-time scaling capability \citep{he2025skywork}.

\begin{figure}[th]
    \centering
  \begin{subfigure}[t]{0.45\linewidth}
        \centering
        \includegraphics[width=0.99\linewidth]{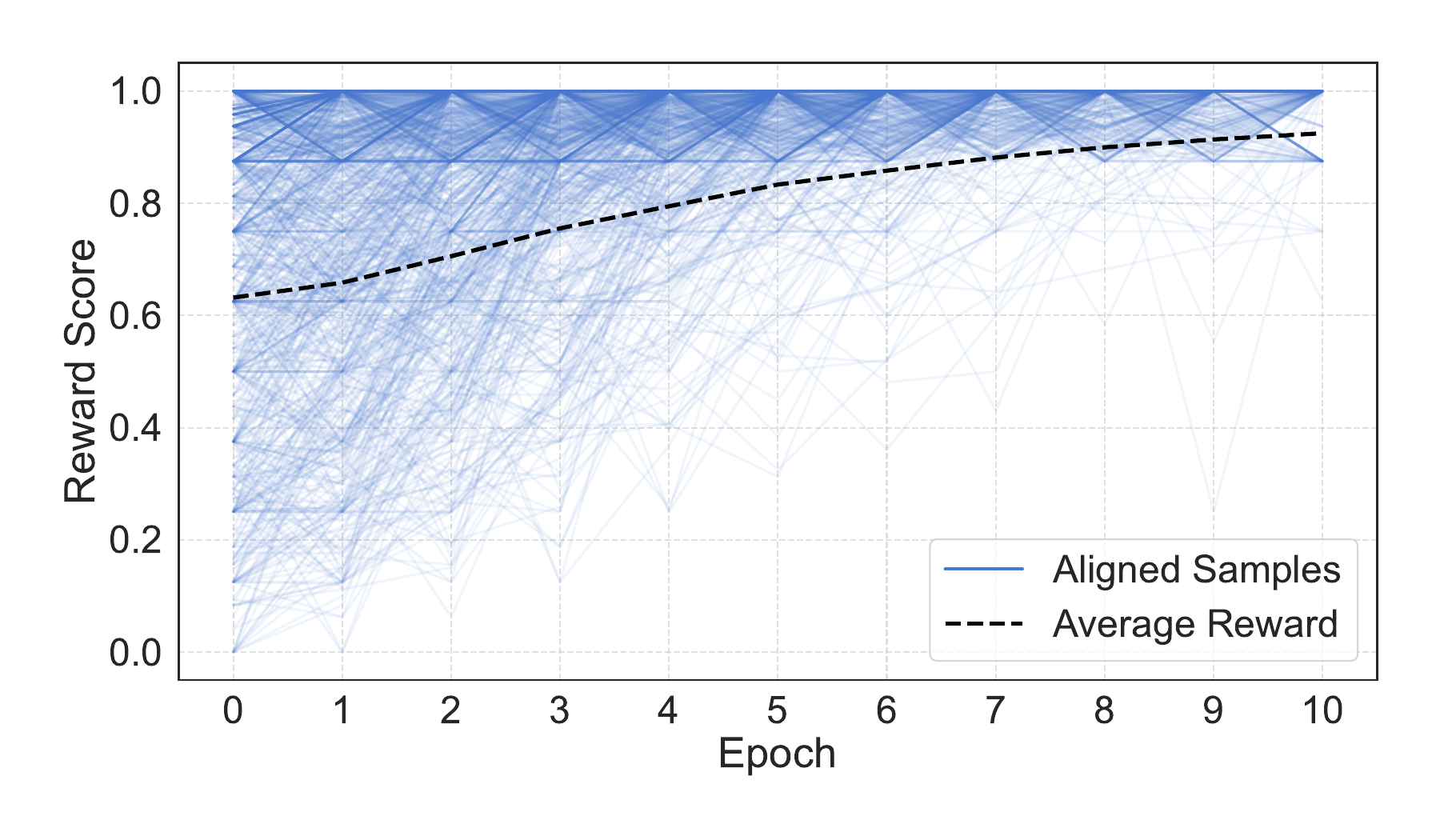}
        \caption{Retained: with aligned samples}
        \label{fig:appx2a}
    \end{subfigure}
    % \hfill
    \begin{subfigure}[t]{0.45\linewidth}
        \centering
        \includegraphics[width=0.99\linewidth]{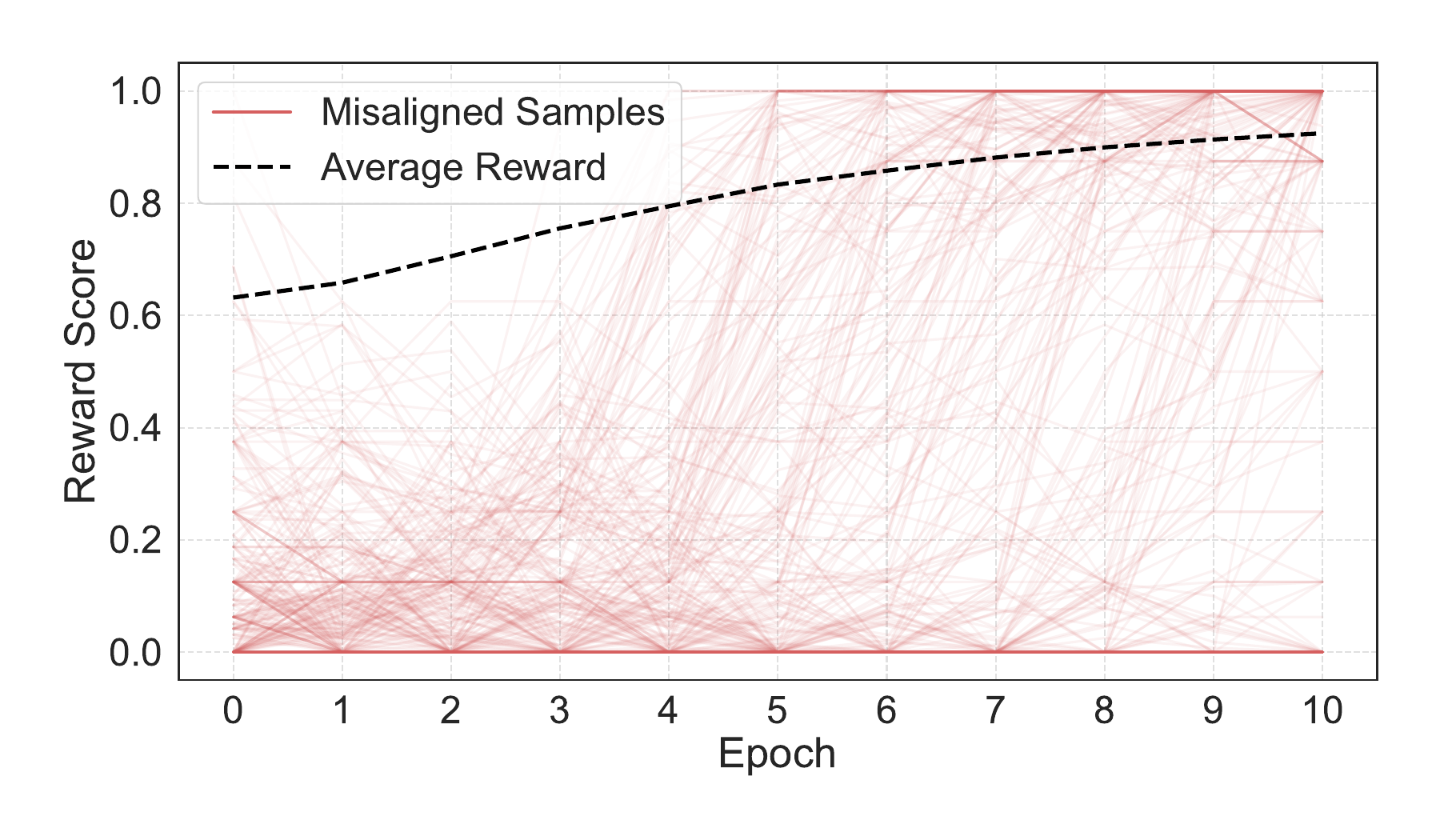}
        \caption{Removed: with misaligned samples}
        \label{fig:appx2b}
    \end{subfigure}
  \caption{
    We retain aligned samples (i.e., those with low variance (a) ) and remove misaligned samples (i.e., those with high variance (b)).
}
  \label{fig:rl_compare}
\end{figure}

\textbf{RL Data Refine Details.}
In the second phase of data processing, we prioritize training samples by their alignment with model learning trajectories, measured through the variance of reward scores from the mean, lower variance indicates better alignment.
\text{Better alignment corresponds to lower variance of reward scores, defined as:}
\[
\text{Var}(r) = \frac{1}{n-1} \sum_{i=1}^{n} (r_i - \mu_r)^2, \quad \mu_r = \frac{1}{n} \sum_{i=1}^{n} r_i
\]
\text{where lower } \text{Var}(r) \text{ indicates better alignment.}
This sampling result is illustrated in Figure \ref{fig:rl_compare}.
From the figure, we observe that the average reward ranges between 0.7 and 0.9, with aligned samples showing higher scores in the upper-left region and misaligned samples displaying lower scores.
To align the dataset size with that of SFT (\~8K samples), we rank examples by reward variance and filter out the bottom 55\%, a simple yet effective heuristic to retain higher-signal instances while controlling scale.

\begin{figure}[th]
    \centering
  \includegraphics[width=0.9\linewidth]{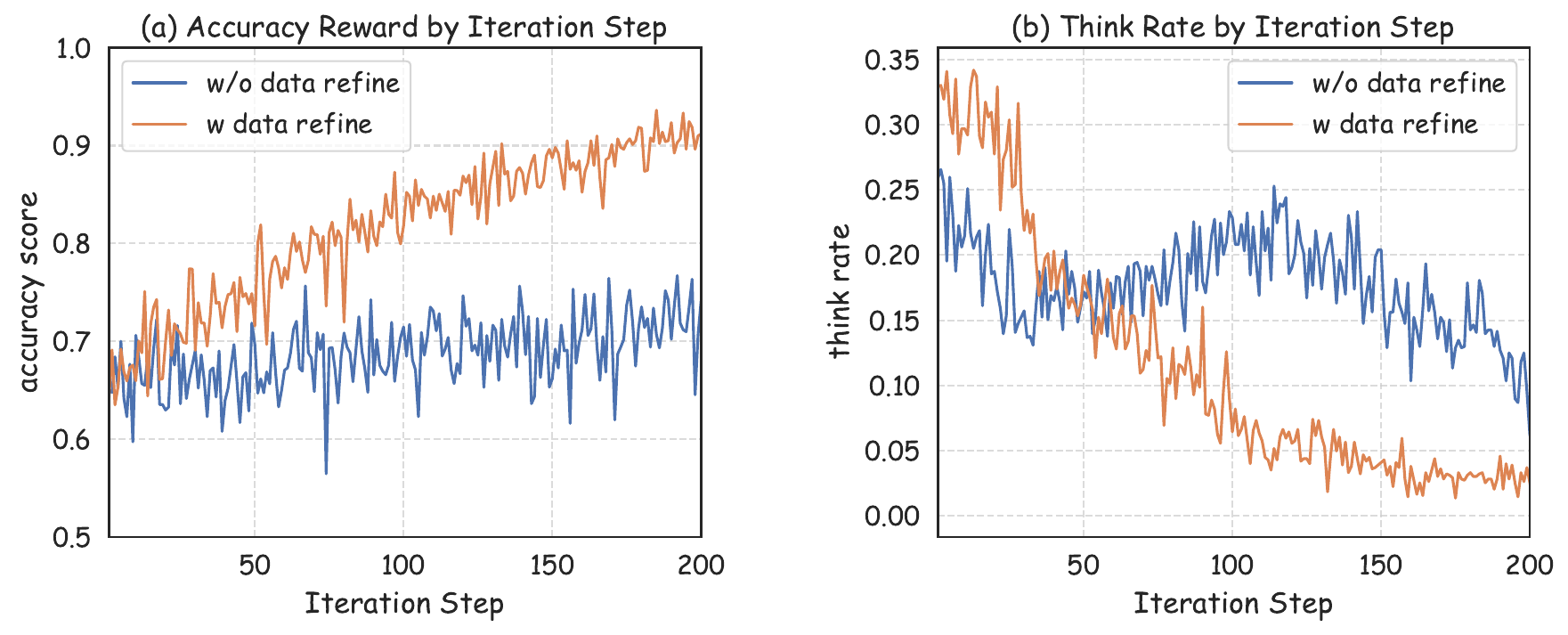}
  \caption{
  RL training processes (with and without data refinement) are shown, along with accuracy scores and thinking rates.
}
  \label{fig:data_compare}
\end{figure}

\textbf{Effectiveness of Data Refinement}.
We experimentally verified its effectiveness. After warm-up SFT, Figure ~\ref{fig:data_compare} shows GRPO training processes with and without data refinement. Results indicate data refinement increases accuracy reward score by +15\%, reduces training fluctuation variance, and enhances stability. 
Additionally, the model's thinking rate converged to a lower level, indicating improved memory capacity. 
BFCL evaluation results show GRPO with data refinement reached 66.82\%, versus 60.78\% without, an improvement of +6.04\%. 
These enhancements are attributed to data refinement filtering substantial noise while retaining high-contribution samples.

% scores/think ratio, long/short len, long/short entropy
\section{Visualization of Training Dynamics}
\label{sec:visualization}
\begin{figure}[th]
    \centering
  \includegraphics[width=0.98\linewidth]{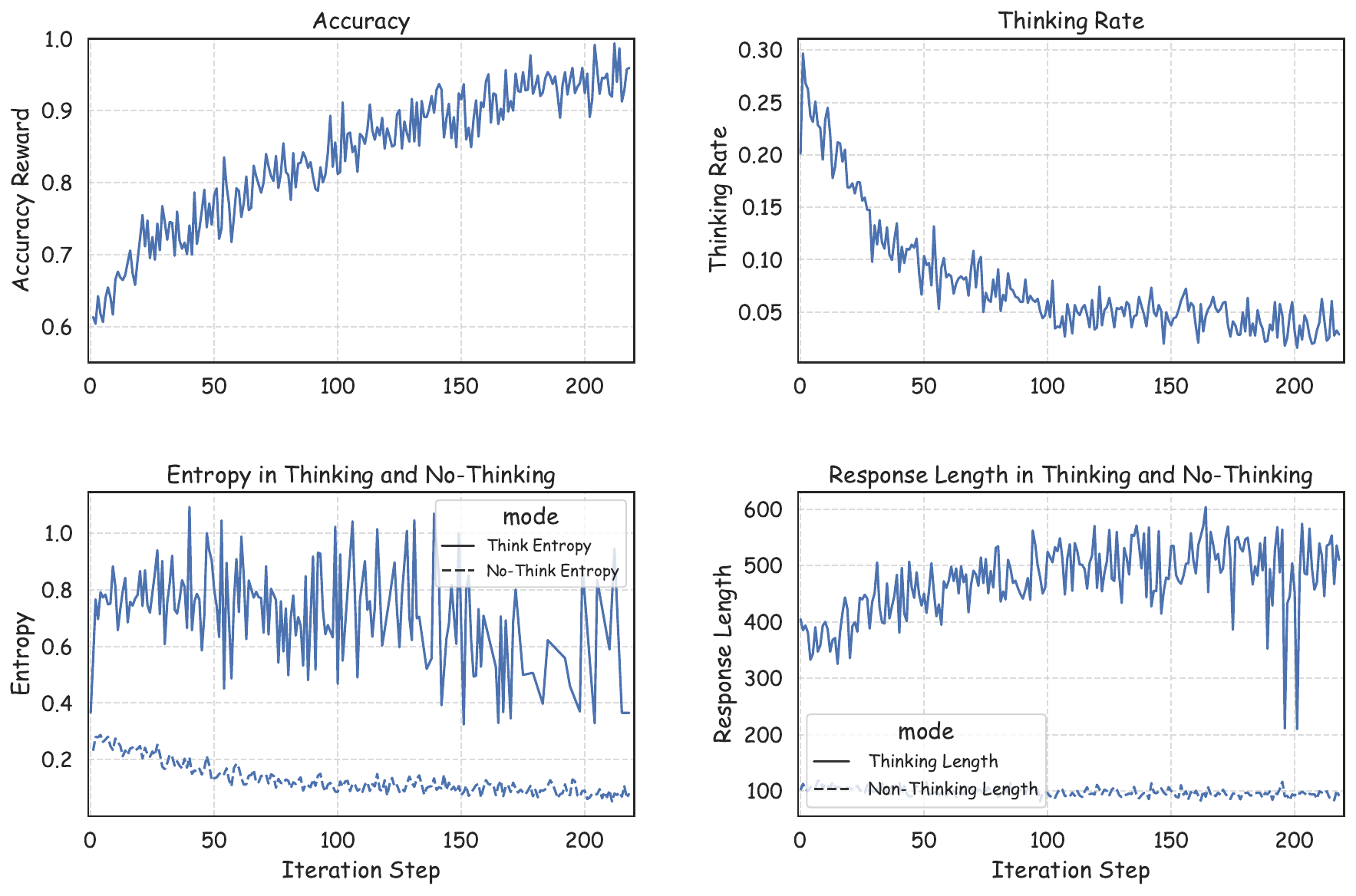}
  \caption{
    The visualization of training dynamics.
    }
  \label{fig:exp5}
\end{figure}

To demonstrate auto-scaling effects, we visualized the training process (Figure~\ref{fig:exp5}).
As training progressed, accuracy improved while the thinking rate gradually decreased to 5\%, indicating fewer problems required long reasoning, suggesting enhanced intrinsic tool-using capabilities.
Additionally, response length and entropy achieved decoupled control: think mode enabled 500\% longer reasoning trajectories than no-think mode, with corresponding higher actor entropy reflecting greater exploration tendency.
These visualizations confirm that training enhanced tool-using abilities and successfully enabled auto test-time scaling based on problem difficulty and model proficiency.

% Generalizability analysis, Backbone Model Analysis.\
% \subsubsection{Generalizability Analysis}
% \label{sec:model_scale}
% We analyze our strategy's performance across diverse backbone models and RLVR algorithms.
% To verify scalability, experiments were conducted using multiple backbone models from Qwen2.5 and LLaMA series. 
% Results in Figure~\ref{fig:exp4a} demonstrate consistent performance gains across different scales and model families, with larger improvements observed in Qwen-series models, indicating better compatibility during test-time scaling. 
% The most significant improvement (16.4\%) was achieved with Qwen2.5-3B-Instruct, confirming generalizability across backbone architectures.

% For RLVR algorithm evaluation (Figure~\ref{fig:exp4b}), we tested GRPO, DAPO (an extension of GRPO \citep{yu2025dapo}), and PPO (with reward models replaced by verifiable reward functions \citep{schulman2017proximal,engstrom2020implementation}). 
% Substantial gains were observed with GRPO (+11.2\%) and DAPO (+8\%), while PPO showed limited improvement (+3.5\%). 
% This highlights GRPO's superior adaptability to reasoning scales in tool learning, attributed to its inter-group policy optimization \citep{qian2025toolrl,muennighoff2025s1}.

\section{Complementary Experiments}
\label{sec:appendix}
\subsection{More Implementation Details}
The experiments were executed using the publicly accessible training framework MindSpeed-RL\footnote{https://gitee.com/ascend/MindSpeed-RL}, an end-to-end reinforcement learning acceleration framework based on the Ascend ecosystem. 
The BFCL is an evolving benchmark and we utilized the version checked out on June 14, 2025.
For the training model, we selected the best performance checkpoint on the valid dataset.
In the Test-time Scale paradigms analysis (\S \ref{sec:pre_study}), we used an instruct model for SFT and a base model for GRPO.
We employ a full-parameter training strategy for all SFT.
In baseline trained on PubTool data, we trained on the complete dataset using specific methods (e.g., SFT, Distilled SFT, and GRPO).
In the PubTool-GRPO training, we adopted the widely used think prompt pattern, which follows the format: \texttt{<think> reasoning process here </think><answer> answer here </answer>}.
Each RL training run for the 7B model completed within 4 hours on a cluster of 32 Ascend 910b NPUs (configured as 4 nodes × 8 NPUs). 
The hyperparameters used are detailed in Table \ref{tab:grpo_config}.

% \noindent
\textbf{Baselines}
% We compare our approach against the following baselines:
(1) \textit{Base Model}: the original model without additional training (e.g., Qwen2.5-series, LLaMA3.1-series).
(2) \textit{SFT-trained Model}: ToolACE-8B (trained on the full ToolACE dataset \citep{liu2024toolace}), xLAM-series (trained on the full xLAM dataset \citep{zhang2024xlam}), and Hammer-series (trained on xLAM with function masking \citep{lin2024hammer}).
(3) \textit{API-based} closed-source frontier models (e.g., GPT-series, Gemini-series).
(4) \textit{RLVR-trained Model}: models trained using GRPO as the RL paradigm, such as QwQ-32B \citep{qwq32b}, Qwen3-series \citep{qwen3technicalreport}, Tool-N1 series (single-turn tool-use models trained on mixed ToolACE and xLAM data \citep{zhang2025nemotron}), and ToolRL(trained in subset of mixed ToolACE and xLAM data \citep{qian2025toolrl}).
Given that we are the first to investigate adaptive reasoning in agentic tool use, there is a significant lack of prior works on this domain, we compare our method with recent adaptive reasoning strategies adapted from other domains (e.g., math). We select two representative methods for comparison: 
(1) Thinkless \citep{fang2025thinkless}: Achieves adaptive reasoning via a control token loss.
(2) Adactrl \citep{huang2025adactrl}: The model self-assesses problem difficulty during the RLVR rollout to facilitate adaptive reasoning.
We replicated both methods using the same dataset (\textit{PubTool}) and the same base model (Qwen2.5-7B-Instruct). 

\begin{table}[th]
    \centering
    \small
    \begin{tabular}{lc|lc}
        \toprule
        Hyperparameter & Value & Hyperparameter & Value \\
        \midrule
        \multicolumn{2}{c|}{\textbf{Data Configuration}} & \multicolumn{2}{c}{\textbf{RL Optimization}} \\ \midrule
        Global Batch Size & 128 & Learning Rate & 1e-6 \\
        Max Prompt Length & 12000 & LR Decay Style & constant \\
        Max Response Length & 2048 & Mini Batch Size & 128 \\
        Micro Batch Size & 4 & KL Loss Used & False  \\
        Train Steps & 200 &  &  \\   \midrule
        \multicolumn{2}{c|}{\textbf{Rollout Configuration}} & \multicolumn{2}{c}{\textbf{Entropy Constraints}} \\ \midrule
        Rollout Name & vllm & Clip Higher $\epsilon$ & 0.28 \\
        GPU Memory Utilization & 0.5 & Think Target Entropy $H_l$ & 0.2 \\
        Number of Rollouts & 8 & No-Think Target Entropy $H_s$ & 0.1 \\ 
        Temperature & 1.0 & Init Adaptive Coefficient $\beta_l$ & 0.1 \\
        Tensor Model Parallel Size & 1 & Fixed Coefficient $\beta_s$ & 0.1 \\
        Top\_P & 1.0 & & \\
        \bottomrule
    \end{tabular}
    \caption{The configurations for RL training with GRPO.}
    \label{tab:grpo_config}
\end{table}

% \section{Hyperparameter Analysis}
% Model performance appears sensitive to target entropies and the initial choice of penalty coefficient $\beta$.
% To identify a suitable target entropy for entropy constraints, we conducted a hyperparameter analysis. 
% We tested values ranging from 0 to 1.0, and their validation set performance is reported in the Figure.
% no-think ent: 0.1 0.2 0.5, think 0.1 0.2 0.5
% beta_s=init beta_l: 0 0.1 0.01 0.001,
% todo
% 拆线图 ，两个reward score and entropy 变化 , 二维变量，考虑使用热图。
% todo 对tgt-ent的值进行消融，对比分析各个0.05, 0.1, 0.2, 0.5 各个值的影响

\subsection{Hyperparameter Sensitivity Analysis}
\label{app:hyperparameter_sensitivity}
In this section, we conduct a hyperparameter sensitivity analysis to explore the effects of key parameters, including the target entropies ($H_s$ and $H_l$) and the entropy penalties ($\beta_s$ and initial $\beta_l$). We evaluated the performance on the BFCL benchmark, reporting the Overall Accuracy.

\textbf{Sensitivity to Target Entropy}
Table~\ref{tab:sensitivity_entropy} presents the sensitivity analysis with respect to the target entropy values. The results indicate that model performance is sensitive to the specific target entropy settings. The optimal performance is achieved when the targets are set around $H_s = 0.1$ and $H_l = 0.2$.

\begin{table}[h]
    \centering
    \begin{tabular}{lccc}
        \toprule
        \textbf{$H_s \setminus H_l$} & \textbf{0.1} & \textbf{0.2} & \textbf{0.5} \\
        \midrule
        \textbf{0.1} & 69.7 & 70.1 & 68.5 \\
        \textbf{0.2} & - & 67.7 & 66.3 \\
        \textbf{0.5} & - & - & 62.1 \\
        \bottomrule
    \end{tabular}
    \caption{Sensitivity analysis in target entropy on the BFCL benchmark. Values represent Overall Accuracy.}
    \label{tab:sensitivity_entropy}
\end{table}

\textbf{Sensitivity to Entropy Penalty}
Table~\ref{tab:sensitivity_penalty} demonstrates the sensitivity analysis regarding the entropy penalty coefficients. The model exhibits considerable stability with respect to $\beta_l$, while lower values of $\beta_s$ generally yield better performance.

\begin{table}[h]
    \centering
    \begin{tabular}{lcccc}
        \toprule
        \textbf{$\beta_s \setminus \beta_l$} & \textbf{0} & \textbf{e-1} & \textbf{e-2} & \textbf{e-3} \\
        \midrule
        \textbf{0} & 70.7 & 68.4 & 71.4 & 70.6 \\
        \textbf{e-1} & 68.8 & 70.1 & 69.5 & 68.7 \\
        \textbf{e-2} & - & - & 69.3 & - \\
        \textbf{e-3} & - & - & - & 70.44 \\
        \bottomrule
    \end{tabular}
    \caption{Sensitivity analysis in entropy penalty on the BFCL benchmark. Values represent Overall Accuracy.}
    \label{tab:sensitivity_penalty}
\end{table}

\subsection{Generalization Across Model Architectures and Sizes}
\label{app:generalization}
To verify the robustness and generalizability of our proposed method, AutoTool, we conducted extensive evaluations across diverse model architectures and parameter scales. 
Specifically, we utilized the Llama 3.2-inst and Qwen 2.5-inst series, covering a wide range of sizes from 1.5B to 32B parameters.

\begin{figure}[h]
    \centering
    \includegraphics[width=0.6\columnwidth]{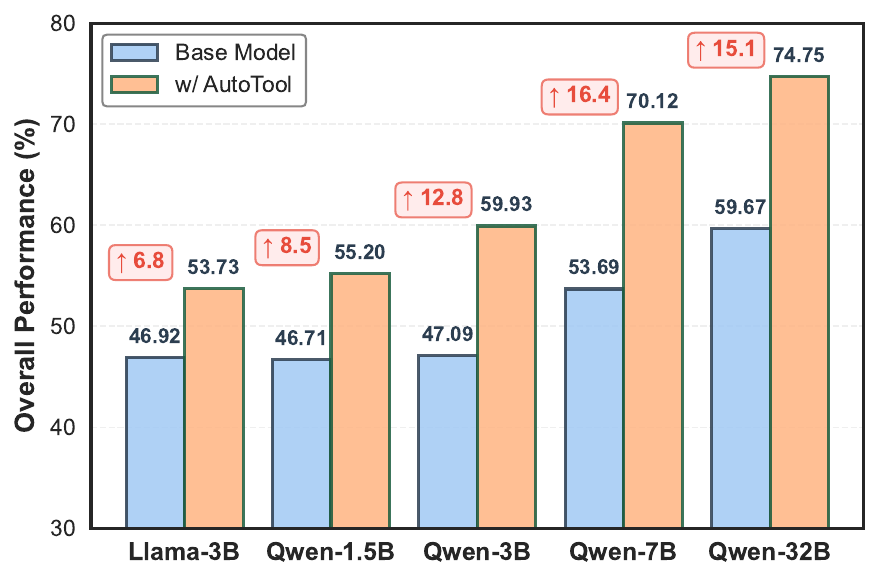} 
    \caption{Performance comparison across different model architectures and sizes on BFCL benchmark. }
    \label{fig:model_comparison}
\end{figure}

Figure~\ref{fig:model_comparison} illustrates the Overall performance on BFCL for both the base models and our method. 
The visual comparison demonstrates that AutoTool consistently improves performance across all tested configurations.

\paragraph{Key Observations:}
\begin{itemize}
    \item \textbf{Substantial Gains in Qwen Series:} The Qwen models exhibited the most significant improvements. As shown in the figure, models in the 3B to 32B range achieved accuracy increases of over 12 points. For instance, the Qwen2.5-7B model saw a remarkable lift from 53.69 to 70.12.
    \item \textbf{Effectiveness on Compact Models:} Even the smallest tested model (1.5B) demonstrated a substantial performance lift (approx. 8.5 points), confirming the method's efficacy for resource-constrained scenarios.
    \item \textbf{Consistency:} The Llama 3.2 architecture also benefited from our approach, with the 3B model improving from 46.92 to 53.73.
\end{itemize}

These findings confirm the generalizability and effectiveness of AutoTool across diverse model landscapes, validating that the adaptive mechanism is not limited to a specific architecture.

\section{Prompt Design for Auto Think}
\label{sec:auto_prompt}
\begin{figure}[t]
    \centering
  \begin{subfigure}[t]{0.45\linewidth}
        \centering
        \includegraphics[width=0.8\linewidth]{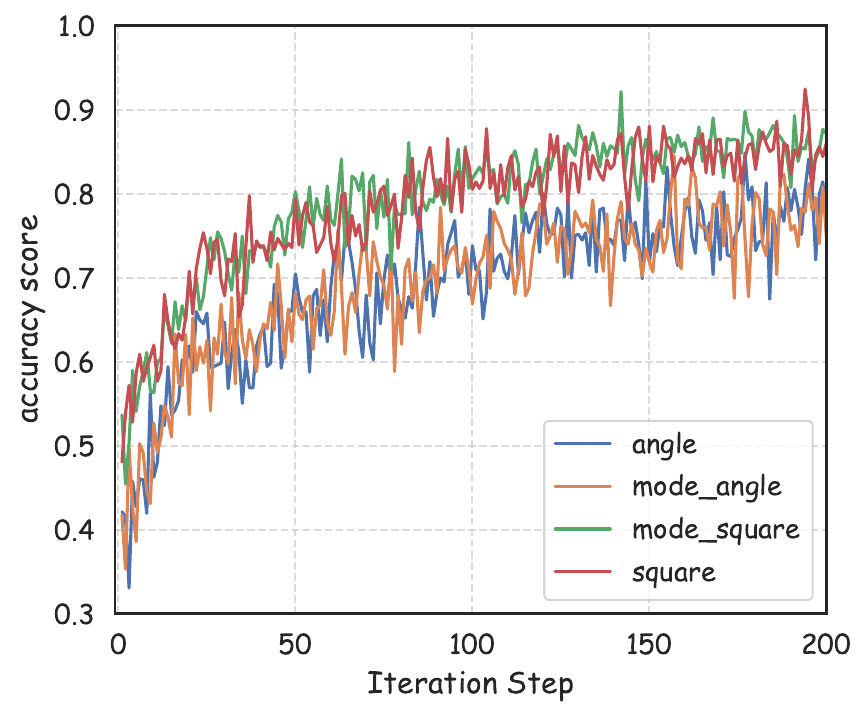}
        \caption{Accuracy Reward by Iteration Step}
        \label{fig:apdx2a}
    \end{subfigure}
    % \hfill
    \begin{subfigure}[t]{0.45\linewidth}
        \centering
        \includegraphics[width=0.8\linewidth]{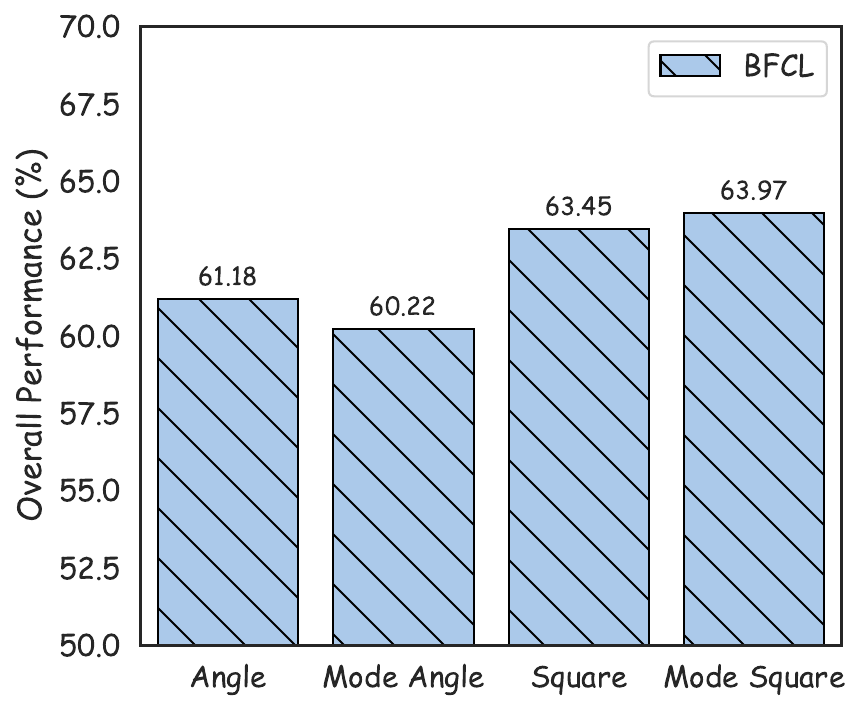}
        \caption{Performance Comparison of Four Prompt Design}
        \label{fig:apdx2b}
    \end{subfigure}
  \caption{
   Visualization of training processes and evaluation results for four prompt designs.
}
  \label{fig:data_refine}
\end{figure}

To explore a suitable prompt design for Auto Think, we conducted a preliminary analysis of the four kind of prompts listed below:
\begin{itemize}
    \item Controlled reasoning mode with square tags: 
    \textbf{[mode]no\_think[/mode] [no\_think] \textbackslash n [/no\_think] [tool\_call] tool calls here [/tool\_call]}
    \item Uncontrolled reasoning mode with square tags: 
    \textbf{[no\_think]\textbackslash n[/no\_think] [tool\_call] tool\ calls here [/tool\_call]}
    \item Controlled reasoning mode with angle tags: \textbf{<mode>no\_think</mode> <no\_think> \textbackslash n </no\_think> <tool\_call> tool calls here </tool\_call>}
    \item Uncontrolled reasoning mode with angle tags: \textbf{<no\_think>\textbackslash n</no\_think> <tool\_call> tool calls here </tool\_call>}
\end{itemize}
We trained the model starting from Qwen2.5-7B-Instruct using the original GRPO algorithm with PubTool LRL data.
Their training processes and evaluation results are presented in Figure \ref{fig:data_refine}. 
From the results, two key observations emerge:
(1) Square tags (\texttt{[]}) exhibit better adaptability than angle tags (\texttt{<>}). 
This may be because the model used angle tags for segmentation in the pre-training phase, reusing these tags for a different purpose (reasoning mode control) is likely to cause signal interference.
(2) Additionally, the explicit "reasoning mode" prefix does not obviously affect performance.
The evaluation results show that the controlled reasoning mode with square tags achieves the best performance; thus, we adopt this prompt design for auto scaling.

% \section{Limitation and Future Work}
% mode size 分析, 不同size 或series模型上是否有效果 
% 其他任务泛化性分析问题, 方法在复杂推理，比如数学上是否也能同等的效果？
% 其他RL算法上的效果，比如 PPO\citep{song2025r1}, DAPO\citep{yu2025dapo}
\section{Limitation and Future Work}
Despite promising results in tool-use scenarios, our method has latent concerns to clear. 
First, we only tested it on a specific model size, future work should verify its scalability across different model scales (e.g., 3B, 13B, 32B parameters) and architecture series, e.g., LLaMA-series.
Second, our method's generalizability beyond tool-use tasks is unproven. 
It is valuable to evaluate its performance on other complex reasoning tasks (e.g., mathematics, logical deduction) to confirm if it can similarly enhance reasoning steps or reduce computational costs. 
Third, our method currently relies on a specific RL algorithm. 
Future research should test its compatibility with other RL algorithms (e.g., PPO \citep{schulman2017proximal,engstrom2020implementation} and DAPO \citep{yu2025dapo}) to verify if the decoupled entropy constraint strategy is effective across different algorithmic paradigms. 
We will address these limitations in future work.

% \begin{figure*}[t]
%     \centering
%     \small
%       \begin{subfigure}[t]{0.99\linewidth}
%         \centering
%         \includegraphics[width=0.99\linewidth]{./pre-exp4_b.pdf}
%         \caption{RL with entropy constraint}
%         \label{fig:exp1a}
%     \end{subfigure}
%     \vspace{1em}
%     \begin{subfigure}[t]{0.99\linewidth}
%         \centering
%         \includegraphics[width=0.99\linewidth]{./pre-exp4_a.pdf}
%         \caption{RL with length penalty}
%         \label{fig:exp1b}
%     \end{subfigure}
%   \caption{
%   Visualizing the training dynamics with entropy constraint (a) and length penalty (b).
%   % todo 介绍结果效果如何
%   }
%   \label{fig:pre-exp3}
% \end{figure*}

\begin{figure*}[th]
    \centering
    \begin{tcolorbox}[
        colback=blue!5!white, 
        colframe=black!80!white, 
        % coltitle=white,                % 白色标题文字
        title= The Full System Prompt for Automatic Think in RL, 
        fonttitle=\bfseries,
        width=0.95\textwidth, 
        left=4mm, right=4mm,
        top=3mm, bottom=3mm,
        center title,
        % arc=4pt,
        arc=6pt,                       % 稍大的圆角更显柔和
        boxrule=1.2pt,                 % 略微加粗边框
    ]
\small
\noindent
You are an advanced function composition agent. Your goal is to solve user queries efficiently. \\
In the multi-turn dialogue loop: the interaction is a cycle: You  \textcolor{blue!90!black}{[tool\_call]}, you receive a \textcolor{blue!90!black}{[tool\_response]}, and you MUST use that new information to plan your next step. \\

\textbf{\# Tools} \\
You are provided with function signatures within \textcolor{blue!90!black}{[tools]} and \textcolor{blue!90!black}{[/tools]} tags: \\
\textcolor{blue!90!black}{[tools] \{functions\} [/tools]} \\

\textbf{\# Action Phase} \\
1. {Choose an Action Mode}: For every turn, you MUST start your response by choosing an action mode ({think} vs {no\_think}) based on the task's complexity. \\
\textcolor{gray!5!white}{no} - {think}: Use for complex reasoning. Enclose your detailed thought process within \textcolor{blue!90!black}{[think]} and \textcolor{blue!90!black}{[/think]} tags. \\
\textcolor{gray!5!white}{no} - {no\_think}: Use for simple, straightforward tasks. You MUST use an empty, self-closing block: \textcolor{blue!90!black}{[no\_think]\textbackslash n[/no\_think]}. \\
    Your response {MUST} begin by enclosing the selected mode name within \textcolor{blue!90!black}{[mode]} and \textcolor{blue!90!black}{[/mode]} tags. \\

2. {Decide on an Action Path}: After that, you must choose ONE of the following two paths: \\

\textbf{\#\# Path A: Call Functions} \\
{WHEN}: The user's intent is tool-related and you have all required functions and parameters. \\
The {tool\_calls} field is a JSON object with function names and arguments within \textcolor{blue!90!black}{[tool\_call]} and \textcolor{blue!90!black}{[/tool\_call]} XML tags. i.e., {
{[tool\_call] [\{"name": <function-name>, "arguments": <args-json-object>\}, \{"name": <function-name2>, "arguments": <args-json-object2>\}, ...] [/tool\_call]} }\\

{EXAMPLE}: \\
\textcolor{orange!90!black}{[mode]no\_think[/mode] [no\_think]\textbackslash n[/no\_think] [tool\_call] tool calls here [/tool\_call]} \\
{EXAMPLE}: \\
\textcolor{orange!90!black}{[mode]think[/mode] [think] reasoning process here [/think] [tool\_call] tool calls here [/tool\_call]} \\

\textbf{\#\# Path B: Respond Directly to the User} \\
{WHEN}: You need to provide a natural language text response. This happens in three main scenarios: \\
(1) After receiving tool execution feedback enclosed within \textcolor{blue!90!black}{[tool\_response]} and \textcolor{blue!90!black}{[/tool\_response]} tags, continue to respond to user queries based on this feedback. \\
(2) The user's query is general conversation and not related to any tool. \\
(3) Ask for more information if the given conversational context lacks the required functions or parameters. \\

{EXAMPLE}: \\
\textcolor{orange!90!black}{[mode]no\_think[/mode] [no\_think]\textbackslash n[/no\_think] natural language sentences you talk with user} \\
{EXAMPLE}: \\
\textcolor{orange!90!black}{[mode]think[/mode] [think] reasoning process here [/think] natural language sentences you talk with user}

    \end{tcolorbox}
    \caption{System Prompt Design for Automatic Scaling Tool-Use in Multi-Turn Dialogue.}
    \label{fig:sys_prompt}
\end{figure*}

\end{document}